\documentclass[a4paper,10pt]{extarticle}
\usepackage{a4wide}
\usepackage{times}
\usepackage[margin=0.65in]{geometry}
\usepackage[utf8]{inputenc} 
\usepackage[T1]{fontenc}    
\usepackage{hyperref}       
\usepackage{url}            
\usepackage{booktabs}       
\usepackage{amsfonts}       
\usepackage{nicefrac}       
\usepackage{microtype}     
\usepackage{amsmath,amssymb}
\usepackage{multirow}
\usepackage{color}
\usepackage{framed}
\usepackage{graphicx}
\usepackage{placeins}
\usepackage{soul}
\usepackage{cleveref}

\makeatother

\def\Ac{\mathcal A}
\def\Bc{\mathcal B}
\def\Ec{\mathcal E}

\makeatletter
\renewcommand{\paragraph}{%
  \@startsection{paragraph}{4}%
  {\z@}{0em}{-0.5em}%
  {\normalfont\normalsize\bfseries}%
}

\title{Powers of layers for image-to-image translation}

\author{Hugo Touvron $^{1,2}$ 
\hspace{0.32cm} 
Matthijs Douze $^1$ 
\hspace{0.32cm} 
Matthieu Cord $^2$
\hspace{0.32cm} 
Herv\'e J\'egou $^1$
\\
~\\
\scalebox{1.}{$^1$ Facebook AI Research \hspace{0.6cm} $^2$ Sorbonne University}
\date{}
}

\begin{document}

\maketitle

\begin{abstract}
We propose a simple architecture to address unpaired image-to-image  translation tasks: style or class transfer, denoising, deblurring, deblocking, etc. 
We start from an image autoencoder architecture with fixed weights. 
For each task we learn a residual block operating in the latent space, which is iteratively called until the target domain is reached. 
A specific training schedule is required to alleviate the exponentiation effect of the iterations. 
At test time, it offers several advantages: the number of weight parameters is limited and the compositional design allows one to modulate the strength of the transformation with the number of iterations. 
This is useful, for instance, when the type or amount of noise to suppress is not known in advance.  
Experimentally, we provide proofs of concepts showing the interest of our method for many transformations. 
The performance of our model is comparable or better than CycleGAN with significantly fewer parameters.
\end{abstract}

\section{Introduction}\label{sec:introduction}

Neural networks define arbitrarily complex functions involved in discriminative or generative tasks by stacking layers, as supported by the universal approximation theorem~\cite{hornik1989multilayer,montufar2014universal,goodfellow2016deep}.
More precisely, the theorem states that stacking a number of basic blocks can approximate any function with arbitrary precision, provided it has enough hidden units, with mild conditions on the non-linear basic blocks.

Studies on non-linear complex holomorphic functions involved in escape-time fractals showed that iterating simple non-linear functions can also construct arbitrarily complex landscapes~\cite{barnsley1988science}.
These functions are complex in the sense that their iso-surfaces are made arbitrarily large by increasing the number of iterations. 
Yet there is no control on the actual shape of the resulting function. This is why generative fractals remain mathematical curiosities or at best
tools to construct intriguing landscapes. 

Our objective is to combine the expressive power of both  constructions, and study 
the optimization of a function that iterates a single building block in the latent space of an auto-encoder. 
We focus on image translation tasks, that can be trained from either \emph{paired} or \emph{unpaired} data.
In the paired case, pairs of corresponding input and output images are provided during training. It offers a direct supervision, so the best results are usually obtained with these methods~\cite{chen2017fast,wang2018pix2pixHD,park2019SPADE}.%

In this paper we focus on the unpaired case: only two corpora of images are provided, one for the input domain $\Ac$ and the other for output domain $\Bc$. Therefore we do not have access to any parallel data~\cite{conneau2017word}, which is a  realistic scenario in many applications, e.g., image restoration. 
We train a function $f_{\Ac\Bc}:\Ac \rightarrow \Bc$, such that the output $b^*=F(a)$ for $a \in \Ac$ is indiscernible from images of $\Bc$. 

\begin{figure*}[t]
\centering
\includegraphics[width=1\linewidth]{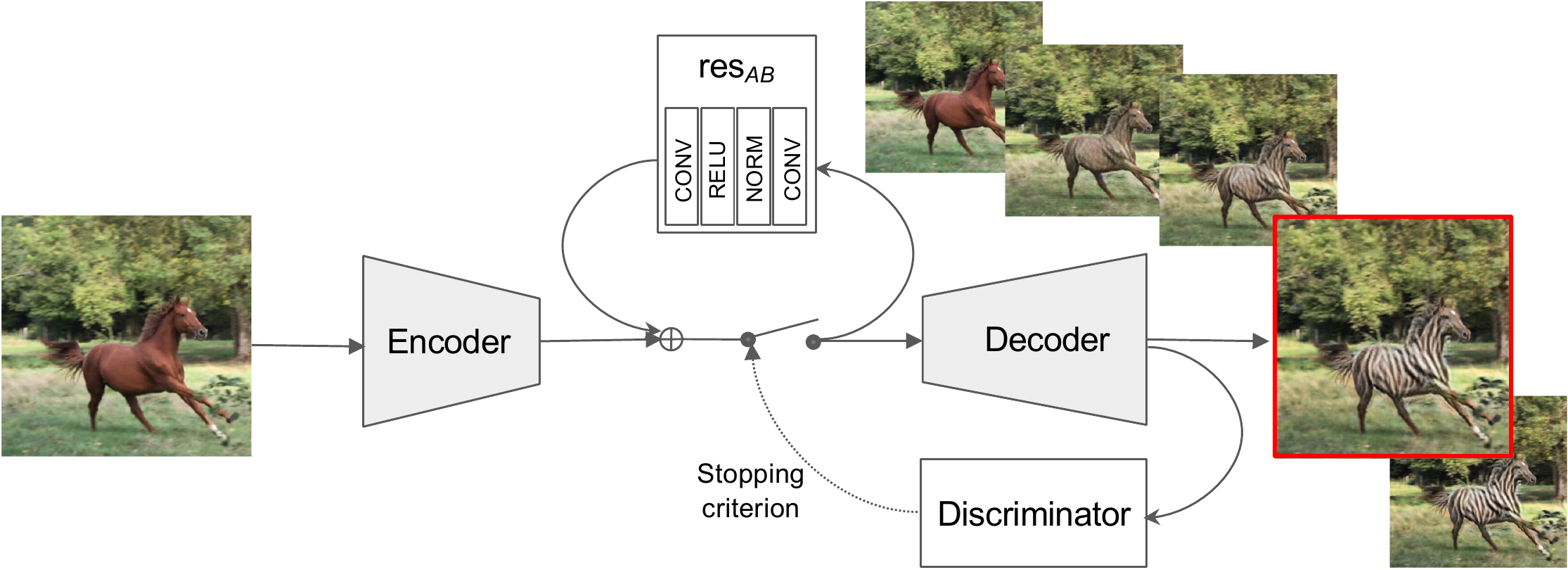}
\vspace{-5pt}
\caption{\label{fig:method} 
Illustration of Powers of layers for a category transfer task. The encoder and decoder are directly borrowed from a vanilla auto-encoder and are not learnable. 
At inference time, we apply a variable number of compositions, producing different images depending on how many times we compose the residual block in the embedding space. 
Depending on the task, we either modulate the transformation and choose the result, or let a discriminator determine when to stop iterating. 
}
\end{figure*}

Our transformation is performed by a single residual block that is composed a variable number of times. 
We obtain this compositional property thanks to a progressive learning scheme that ensures that the output is valid for a large range of iterations.
As a result, we can modulate the strength of the transformation by varying the number of times the transformation is composed. This is of particular interest in image translation tasks such as denoising, where the noise level is unknown at training time, and style transfer, where the user may want to select the best rendering. 
This ``Powers of layers'' (PoL) mechanism is illustrated in Figure~\ref{fig:method} in the category transfer context (horse to zebra).

Our architecture is very simple and only the weights of the residual block differ depending on the task, which makes it suitable to address a large number of tasks with a limited number of parameters. 
This proposal is in sharp contrast with the trend of current state-of-the-art works to specialize the architecture and to increase its complexity and number of parameters~\cite{Fu2019GeometryConsistentGA,viazovetskyi2020stylegan2,choi2019starganv2}. 
Despite its simplicity, our proof of concept exhibits similar or better performance than a vanilla CycleGAN architecture, all things being equal otherwise, for the original set of image-to-image translation tasks proposed in their papers, as well as for denoising, deblurring and deblocking. With significantly fewer parameters and a versatile architecture, we report competitive results confirmed by objective and psycho-visual metrics,  illustrated by visualizations.  

\section{Related work}\label{sec:related}

\paragraph{Generative adversarial networks (GANs)}~\cite{Goodfellow2014GenerativeAN} is a framework where two networks, a generator and a discriminator, are learned together in a zero-sum game fashion. 
The generator learns to produce more and more realistic images wrt. the training dataset with real images.
The discriminator learns to discriminate better and better between real data and generated images. 
GANs are used in many tasks such as domain adaptation, style transfer, inpainting and talking head generation~\cite{Bousmalis2016DomainSN,Karras2019AnalyzingAI,Pathak2016ContextEF,Zakharov2019FewShotAL}. 

\paragraph{Unpaired image-to-image translation} considers the tasks of  transforming an image from a domain $\Ac$ into an image in a domain $\Bc$. 
The training set comprises a sample of images from domains $\Ac$ and $\Bc$, but no pairs of corresponding images. 
A classical approach is to train two generators ($\Ac \rightarrow \Bc$ and $\Bc\rightarrow \Ac$) and two discriminators, one for each domain. 
When there is a shared latent space between the domains, a possible choice is to use a variational auto encoder like in CoGAN~\cite{Liu2016CoupledGA}. 
CycleGAN~\cite{Zhu2017CycleGAN}, DualGAN~\cite{Yi2017DualGAN} and subsequent works~\cite{huang2018munit,liu2019FUNIT,Liu2017UnsupervisedIT,Fu2019GeometryConsistentGA,choi2019starganv2} augment the adversarial loss induced by the discriminators with a cycle consistency constraint to preserve semantic information throughout the domain changes. 
All these variants have architectures roughly similar to CycleGAN: an encoder, a decoder and residual blocks operating on the latent space. 
They also incorporate elements of other networks such as StyleGAN~\cite{Karras2018ASG}.
In our work, we build upon a simplified form of the CycleGAN architecture that generalizes over tasks easily.

\paragraph{High resolution images with GANs. } 
Generating high-resolution images is challenging. 
Until recently, GANs architectures were designed to produce low-resolution images.
Indeed, the memory usage at training time depends heavily on the size of the images. 
The general approach is to produce the target image in a scale pyramid, which outputs results of much finer quality~\cite{karras2017progressive,Karras2018ASG,Brock2018LargeSG,choi2019starganv2,Lin2019COCOGANGB,Karnewar2019MSGGANMG}.
Generating high resolution images makes it possible to get closer to the format of real pictures used in CGI~\cite{Karras2019AnalyzingAI,choi2019starganv2} or medical imaging~\cite{Zbontar2018fastMRIAO}. 

\paragraph{Transformation modulation}
is an interpolation between two image domains.
It is a byproduct of some approaches~\cite{lample2017fader,Brock2018LargeSG,Radford2015UnsupervisedRL}.
For instance, a linear interpolation in latent space~\cite{Brock2018LargeSG,Radford2015UnsupervisedRL} morphs between two images.
Nevertheless, one important limitation is that the starting and ending points must both be known, which is not the case in unpaired learning. 
Other approaches such as the Fader networks~\cite{lample2017fader} or StyleGan2~\cite{viazovetskyi2020stylegan2} act on scalar or boolean attributes that are disentangled in the latent space ({\it eg.,} age for face images, wear glasses or not, etc). 
Nevertheless, this results in complex models, 
for which dataset size and the variability of images strongly impacts the performance: they fail to modulate the transform with small datasets or with large variabilities. 
A comparison of PoL with the Fader network is provided in Appendix~\ref{app:cyclegan} and shows that our approach is more effective. 

\paragraph{Progressive learning and inference time modulation}
are performed in multi-scale methods such as SinGAN~\cite{rottshaham2019singan} and ProgressiveGAN~\cite{karras2017progressive}. 
Progressive learning obtains excellent results for high resolution images where it is more difficult to use classical approaches. 
The training is performed in several steps during which the size of both the images and the network are increased.

The inference time of some architectures can be modulated by stopping the forward pass at some layer~\cite{Huang2016DeepNW,Wu2017BlockDropDI,Veit2017ConvolutionalNW,Figurnov2016SpatiallyAC}.
This differs from our approach, where the number of residual block compositions (``powers'') can be chosen, to shorten the inference. 
A by-product is a reduction of the number of network parameters.

\paragraph{Weight sharing}
is a way of looking at our method, because the same layer is applied several times within the same network.
Recurrent Neural Networks (RNN) are the most classical example weight sharing in a recursive architecture.
Besides RNNs, weight sharing is mainly used for model compression, sequential data and ordinary differential equations (ODE)~\cite{Gao2019CrossDM,Polino2018ModelCV,Han2015DeepCC,Chen2018NeuralOD}. 
A few works~\cite{Jastrzebski2017ResidualCE,Zhang2018RecurrentCA} apply weight sharing to unfold a ResNet and evaluate its performance in classification tasks.
The optimization is inherently difficult, so they use independent batch normalization for each shared layer.
With PoL we observe the same optimization issues, that we solve by a progressive training strategy, see Section~\ref{sec:progressive}.
Recent work~\cite{Jeon2020DifferentiableFI} are interested in the composition of the same block by considering the parallel with the fixed point theorem, nevertheless their application remains to rather simple problems compared to our unpaired image-to-image translation tasks. 

\section{Power of layers}\label{sec:method}

\newcommand\norm[1]{\left\lVert#1\right\rVert}
We adopt the same context as CycleGAN~\cite{Zhu2017CycleGAN} and focus on unpaired image to image translation: the objective is to transform an image from domain $\Ac$ into an image from domain $\Bc$.
In our case the domains can be noise levels, painting styles, blur, JPEG artifacts, or simply object classes that appear in the image.
The training is unpaired: we do not have pairs of corresponding images at training time. 
CycleGAN is simple, adaptable to different tasks and allows a direct comparison in Sections~\ref{sec:analysis} and~\ref{sec:experiments}. 

We learn two generators and two discriminators.
The generator $G_{AB}: I \rightarrow I$ transforms an element of $\Ac$ into an element of $\Bc$, and $G_{\Bc\Ac}$ goes the other way round, $I$ being the fixed-resolution image space.
The discriminators $D_{\Ac}: I \rightarrow [0, 1]$ (resp. $D_{\Bc}$) predicts whether an element belongs to domain $\Ac$ (resp $\Bc$).
We use the same losses as commonly used in unpaired image-to-image translation:

\begin{align}
\mathcal{L}_\mathrm{Total} =  & \ 
\lambda_\mathrm{Adv}\mathcal{L}_\mathrm{Adv} + 
\lambda_\mathrm{Cyc}\mathcal{L}_\mathrm{Cyc} + 
\lambda_\mathrm{Id}\mathcal{L}_\mathrm{Id},    \\
\textrm{\quad \quad where } & \mathcal{L}_\mathrm{Adv}(G_{AB},D_{B})  =
\mathbb{E}_{b\sim \Bc}[\log D_{B}(b)]+ 
\mathbb{E}_{a\sim \Ac}[\log(1- D_{B}(G_{AB}(a)))], \nonumber \\
& \mathcal{L}_\mathrm{Cyc}(G_{AB},G_{BA}) =
\mathbb{E}_{b\sim \Bc}[\norm{G_{AB}(G_{BA}(b))-b}_1]+\mathbb{E}_{a\sim \Ac}[\norm{G_{BA}(G_{AB}(a))-a}_1], \nonumber  \\
& \mathcal{L}_\mathrm{Id}(G_{AB},G_{BA}) =
\mathbb{E}_{b\sim \Bc}[\norm{G_{AB}(b)-b}_2]+\mathbb{E}_{a\sim  \Ac}[\norm{G_{BA}(a)-a}_2]. \nonumber
\end{align}

The Adversarial loss $\mathcal{L}_\mathrm{Adv}(G_{AB},D_{B})$ verifies that the generated images are in the correct domain. The Cycle Consistency loss $ \mathcal{L}_\mathrm{Cyc}(G_{AB},G_{BA})$ ensures a round-trip through the two generators reconstructs the initial image, and the {identity loss} $ \mathcal{L}_\mathrm{Id}(G_{AB},G_{BA})$ penalizes the generators transforming images that are already in their target domain. We keep the same linear combination coefficients as in CycleGAN~\cite{Zhu2017CycleGAN}:
$\lambda_\mathrm{Adv}=1$, $\lambda_\mathrm{Cyc}=10$, $\lambda_\mathrm{Id}=5$.

\newcommand{\Enc}{\mathrm{Enc}}
\newcommand{\Dec}{\mathrm{Dec}}
\newcommand{\res}{\mathrm{res}}

\subsection{Network architecture}
\label{sec:arch}

We start from the CycleGAN architecture~\cite{Zhu2017CycleGAN}. 
The encoder and decoder consist of 2 layers and a residual block. 
The embedding space $\Ec$ of our model is $256 \times 64 \times 64$: its spatial resolution is 1/4 the input image resolution of $256\times256$ and it has 256 channels.
All translation operations take place in the fixed embedding space $\Ec$.
The encoder $\Enc: I \rightarrow \Ec$ produces the embedding and consists of two convolutions. 
The decoder $\Dec: \Ec \rightarrow I$ turns the embedding back to image space and consists of two transposed convolutions.

Note that we will provide the implementation for the sake of reproducibility.

\paragraph{Pre-training of the auto-encoder.}
We train the encoder and decoder of our model 
on a reconstruction task with 6M images randomly drawn from the YFCC100M dataset~\cite{Thomee2016YFCC100MTN} during one epoch, using an $\ell_2$ reconstruction loss in pixel space. 
We use the Adam optimizer with a learning rate of $16 \times 10^{-4}$.
Our data-augmentation consists of an image resizing, a random crop and a random horizontal flip. 
Both the encoder and decoder weights are fixed for all the other tasks, only the residual block is adapted (and the discriminator in case we use it for the stopping criterion). 

\paragraph{The embedding transformer -- single block.}

The transformation between domains is applied is based on a residual block $f_{\Ac\Bc}$, similar to the feed-forward network used in transformers~\cite{Vaswani2017AttentionIA}.
It writes: 
\begin{equation}
    f_{\Ac\Bc}(x) = x + \res_{\Ac\Bc}(x), 
    \forall x \in \Ac.
\end{equation}

\newcommand{\expfactor}{K}

There is a dimensionality expansion factor $\expfactor$ between the two convolutions in the residual block (see Figure~\ref{fig:method}). 
Adjusting $\expfactor$ changes the model's capacity. We adopt the now standard choice of the original transformer paper ($\expfactor=4$). 
The full generator writes
\begin{equation}
    G_{\Ac\Bc}(x) = \Dec(f_{\Ac\Bc}(\Enc(x))), \forall x \in \Ac.
\end{equation}
The other direction, with $f_{\Bc\Ac}$ and $\res_{\Bc\Ac}$, is defined accordingly.

\paragraph{Powers of layers. }

We start from the architecture above and augment its representation capacity. 
There are two standard ways of doing this: (1) augmenting the capacity of the $f_{\Ac\Bc}$ block by increasing $\expfactor$; (2) increasing the depth of the network by chaining several instances of $f_{\Ac\Bc}$, since the intermediate representations are compatible. 

In contrast to these fixed architectures, PoL \emph{iterates} the $f_{\Ac\Bc}$ block $n \geq 1$ times, which amounts to sharing the weights of a deeper network:
\begin{equation}
    G_{\Ac\Bc}(x) = \Dec(f_{\Ac\Bc}^n(\Enc(x))), \forall x \in \Ac.
\end{equation}

\subsection{Optimization in a residuals blocks weight sharing context}

In the following, we drop the $\Ac\Bc$ suffix from $f_{\Ac\Bc}$, since powers of layers operates in the same way on $f_{\Ac\Bc}$ and $f_{\Bc\Ac}$. %
Thus, 
$ f:\Ec \rightarrow \Ec$ is $f(x) = x + \res(x)$. 
The parameters of $f$ are collected in a vector $w$.
The embedding $x\in \Ec$ is 3D activation map, but for the sake of the mathematical derivation we linearize it to a vector.
The partial derivatives of $f$ are $\frac{\partial f}{\partial x} =\frac{\partial \res}{\partial x} + \mathrm{Id}$ and $\frac{\partial f}{\partial w} =\frac{\partial \res}{\partial w}$.
 We compose the $f$ function $n$ times as

\begin{equation}
\frac{\partial f^n}{\partial x}(x)~=~\prod_{i=n-1}^0 
    \frac{\partial f}{\partial x}(f^i(x))
\textrm{\quad and \quad  }
    \frac{\partial f^n}{\partial w}(x)~=~\prod_{i=n-1}^1 
    \frac{\partial f}{\partial x}\left(f^i(x)\right)
    \frac{\partial f}{\partial w}(x). 
\end{equation}

The stability of the SGD optimization depends on the magnitude and conditioning of the matrix $M_n$ defined as:

\begin{equation}
M_n = \prod_{i=n-1}^1 
    \frac{\partial f}{\partial x}\left(f^i(x)\right)
= \prod_{i=n-1}^1 
    \left( 
    \frac{\partial \res}{\partial x}\left(f^i(x)\right)
    + \textrm{Id}
    \right), 
\end{equation}

which is sensitive to initialization during the first optimization epochs.
Indeed, the length of the SGD steps on $w$ depends on the eigenvalues of $M_n$.
When simplifying the basic residual block to a linear transformation $L\in \mathbb{R}^{d\times d}$ (i.e., ignoring the normalization and the ReLU non-linearity), we have $M_n = (L + \mathrm{Id})^{n-1} $.
The eigenvalues of $M_n$ are $(\lambda_i + 1)^{n-1}$, where $\lambda_1, ...,\lambda_d$ are the eigenvalues of $L$. 
At initialization, the components of $L$ are sampled from a random uniform distribution. 
To reduce the magnitude of $\lambda_i$, one option is to make the entries of $L$ small. 
However, to decrease  $(\lambda_i + 1)^{n-1}$ sufficiently, $\lambda_i$ must be so small that it introduces floating-point cancellations when the residual block is added back to the shortcut connection. 
This is why we prefer to adjust $n$, as detailed next.

\subsection{Progressive training}
\label{sec:progressive}

We adopt a progressive learning schedule in a ``warm up'' phase: %
we start the optimization with a single block and add one iteration at every epoch until we reach the required $n$ blocks.
This is possible because the blocks operate in the same embedding space $\Ec$ and because their weights are shared, so all blocks are still in the same learning schedule.
In addition, this approach allows the discriminator to improve progressively during the training. 
For example, in the case of the transformation horse $\rightarrow$ zebra, a slightly whitened horse fools the discriminator at the beginning of the training, but a stronger stripes texture is required later on.

\paragraph{Training for modulation.}
If the network is trained with a fixed number of compositions, the intermediate states do not correspond to modulations of the transformation that ``look right'' (see Appendix~\ref{app:training}).
Therefore, in addition to this scheduled number of iterations during the first $n$ epochs of warm up, we also randomize the number of iterations in subsequent epochs. 
This forces the generator to also produce acceptable intermediate states, and enables modulating the transform.

\paragraph{Stopping criterion at inference time.} 
Each image of domain $\Ac$ is more or less close to domain $\Bc$.
For example, when denoising, the noise level can vary so the denoising strength should adapt to the input. 
Similarly for horse$\rightarrow$zebra: a white horse is closer to a zebra than a brown horse.
Therefore, at inference time, we can adjust the number of compositions as well. 
In particular, for each test image, we select $n$ that best deceives the discriminator, thus effectively adapting the processing to the relative distance to the target domain.  

\section{Analysis}\label{sec:analysis}

\newcommand{\stdminus}[1]{ \scalebox{0.8}{$\pm #1$}}

In this section we study the impact of the main training and design choices and on the performance of powers of layers. 
Appendices~\ref{app:training} and~\ref{app:inference} provide complementary analysis for training- and inference-time choices, respectively. 

For this preliminary analysis, we focus on denoising tasks for which the performance is easily measurable. We add three types of noise to images: 
Gaussian noise, Gaussian blur and JPEG artifacts. 
The noise intensity is quantified by the noise standard deviation, the blur radius and the JPEG quality factor, respectively. 
We generate transformed images and measure how well our method recovers the initial image. 

Note that, in the literature, these tasks are best addressed by providing  (original, noisy) pairs if images. 
Our objective is to remain in a completely unpaired setting during the training phase. 
It corresponds to the case where parallel data is not available (like for the restoration of ancient movies), and also better reflects the situation where the noise strength is not known in advance. 
Therefore, the original image is solely employed to measure the performance.
This provides a more reliable signal than more classical unpaired image-to-image translation evaluations. 
\newcommand{\maxcomp}{n_\mathrm{tr}}
\newcommand{\maxcomptest}{n_\mathrm{te}}

\paragraph{Experimental protocol.}
To train, we sample 800 domain $\Ac$ images from the high-resolution Div2K dataset~\cite{Agustsson_2017_CVPR_Workshops}. 
In the baseline training procedure, the warm up phase starts from a single block and increases the number of compositions at every epoch, until we reach epoch $\maxcomp$. Then we keep the number of compositions fixed. 

We test on the Urban-100~\cite{Huang2015Urban100} dataset.
Unless specified otherwise, we set the number of compositions to $\maxcomptest = \maxcomp$ and
measure the Peak Signal to Noise Ratio (PSNR) of our model on the dataset images, degraded with the same intensity as at training time.
For the JPEG case we use the  Naturalness Image Quality Evaluator metric~\cite{Mittal2013NIQE} (NIQE, lower=better) instead, because it is more sensitive to JPEG artifacts.
NIQE is a perceptual metric that does not take the original image into account.

\begin{table*}
\centering
\scalebox{0.8}
{
\small
\begin{tabular}{|c|cc|cc|}
  \toprule

      & \multicolumn{2}{c}{Gaussian noise (std=30)} & \multicolumn{2}{|c|}{Gaussian blur ($\sigma$=4)}\\
     $\maxcomp$  & POL &  ind & POL &  ind  \\
    \midrule
    1 &  23.3 & \textbf{23.3}&  18.6 & 18.6 \\
    2 &  23.3 & 23.2&  18.5 & 18.6 \\
    4 &  24.4 &  23.2&  19.2 & 19.2 \\
    8 &  23.9 & 22.3 &  19.0 & \textbf{19.3} \\
    12 &  23.9 & 22.5 &  \textbf{19.7} & 18.8 \\
    16 &  23.9 & 22.5&  19.0 & 18.1 \\
    18 &  \textbf{24.2} & \_ &  19.0 & \_  \\
    24  &  23.8 &\_ &  18.6 & \_  \\
    30 &   23.5 & \_ & 19.0&\_  \\
    \bottomrule
\end{tabular}
}
\hfill
\begin{minipage}{0.55\linewidth}
\caption{\label{tab:psnr_urban100_sharing_max_rep}
Denoising: PSNR on Urban-100~\cite{Huang2015Urban100}.
Comparison between Power of layers (POL) and independent (ind) blocks for different maximum number of composition / residual block. Best value for each column  are in \textbf{Bold}.
We could not fit more than 16 independent blocks in memory in our experiments.
 We provide standard deviation and additional results in Appendix~\ref{app:training}.
} \vfill
\end{minipage}
\end{table*}

\paragraph{Block composition or independent successive blocks?}

Table~\ref{tab:psnr_urban100_sharing_max_rep} compares PoL's composition of the same block versus using independent blocks with distinct weights. 
In spite of the much larger capacity offered by independent blocks, the noise reduction operated by Power of layers is stronger. 
Our interpretation is that the model is easier to train. 

\begin{figure*}
\scalebox{0.76}{
\small
\begin{tabular}{c@{}cc@{}}
     & Discriminator of Natural images & Discriminator of Noisy images \\
     \midrule
     \rotatebox{90}{\quad \quad Fixed $\maxcomp=30$} &
      \includegraphics[width=0.40\linewidth]{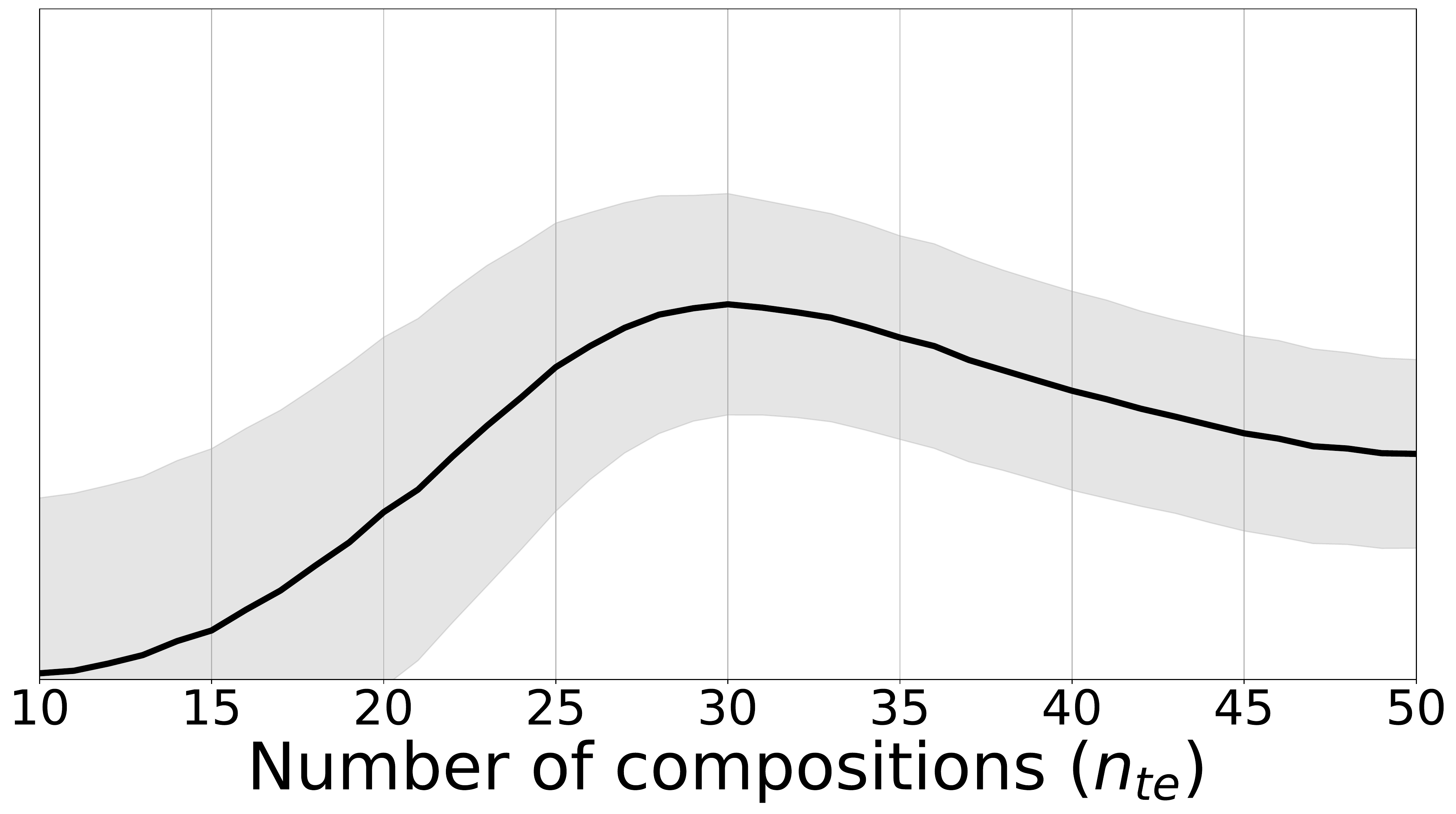}&
    \includegraphics[width=0.40\linewidth]{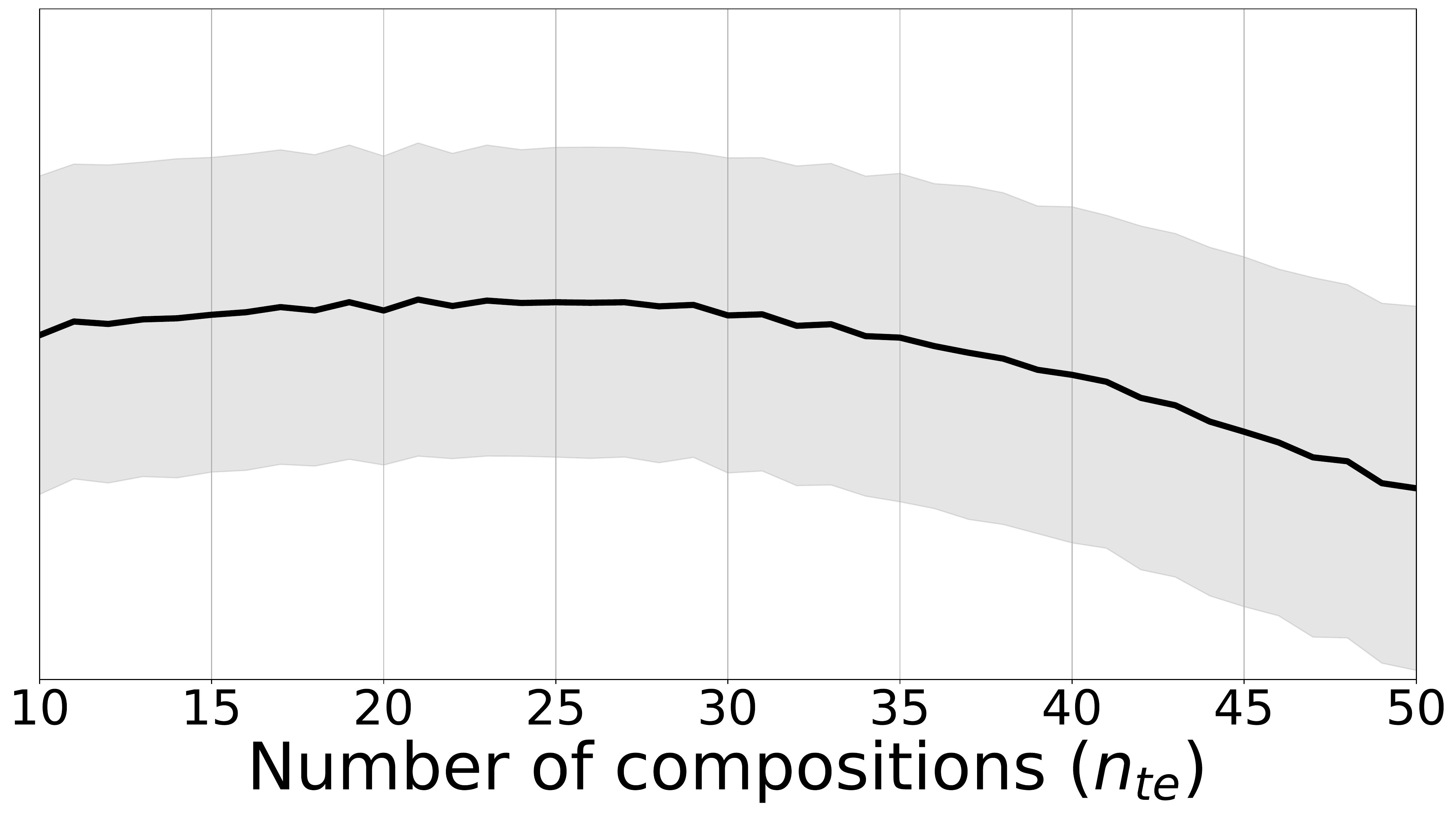}\\
     \midrule
   \rotatebox{90}{Random $\maxcomp \in [\![20, 30]\!]$} &
      \includegraphics[width=0.40\linewidth]{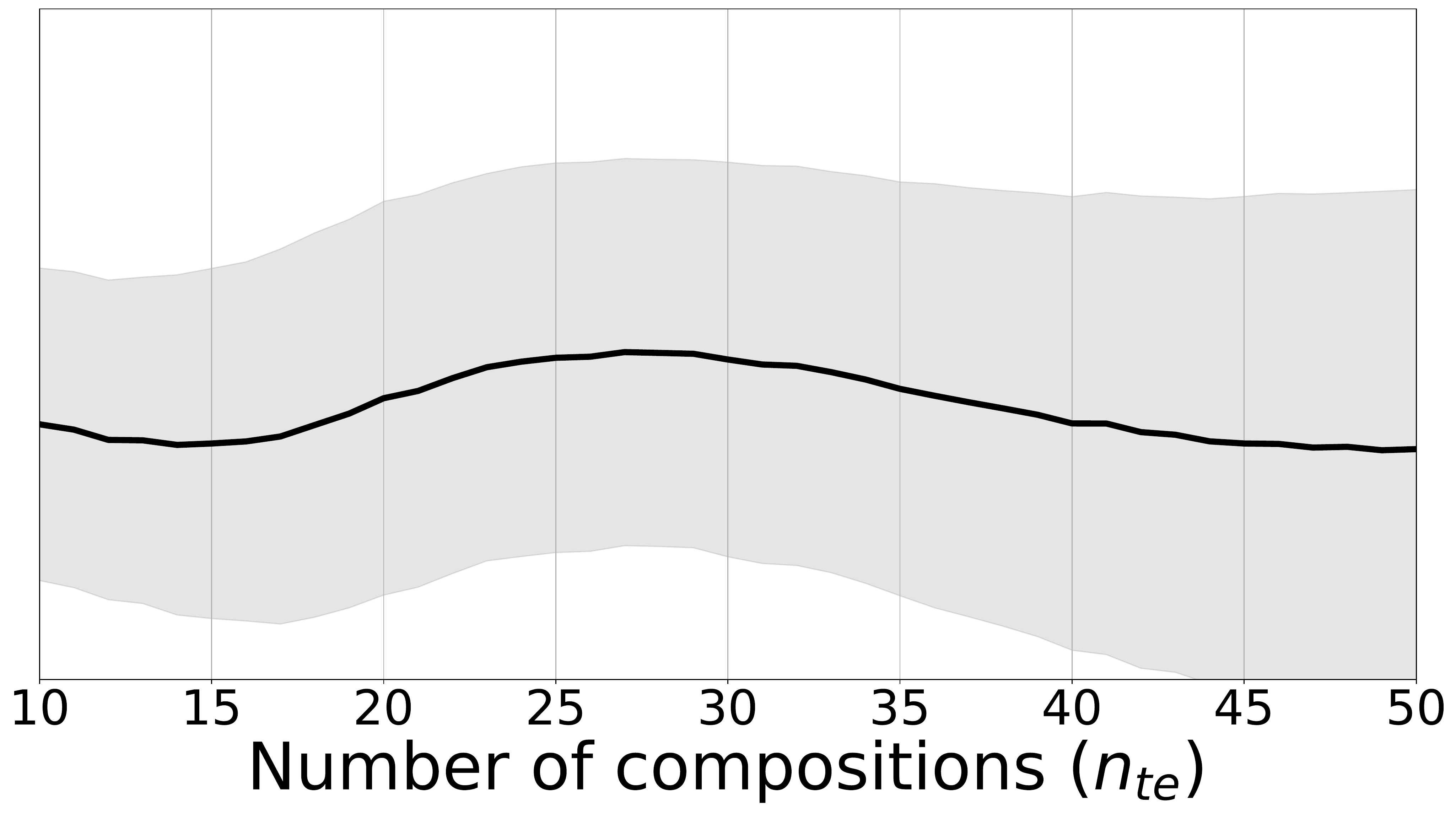}&
      \includegraphics[width=0.40\linewidth]{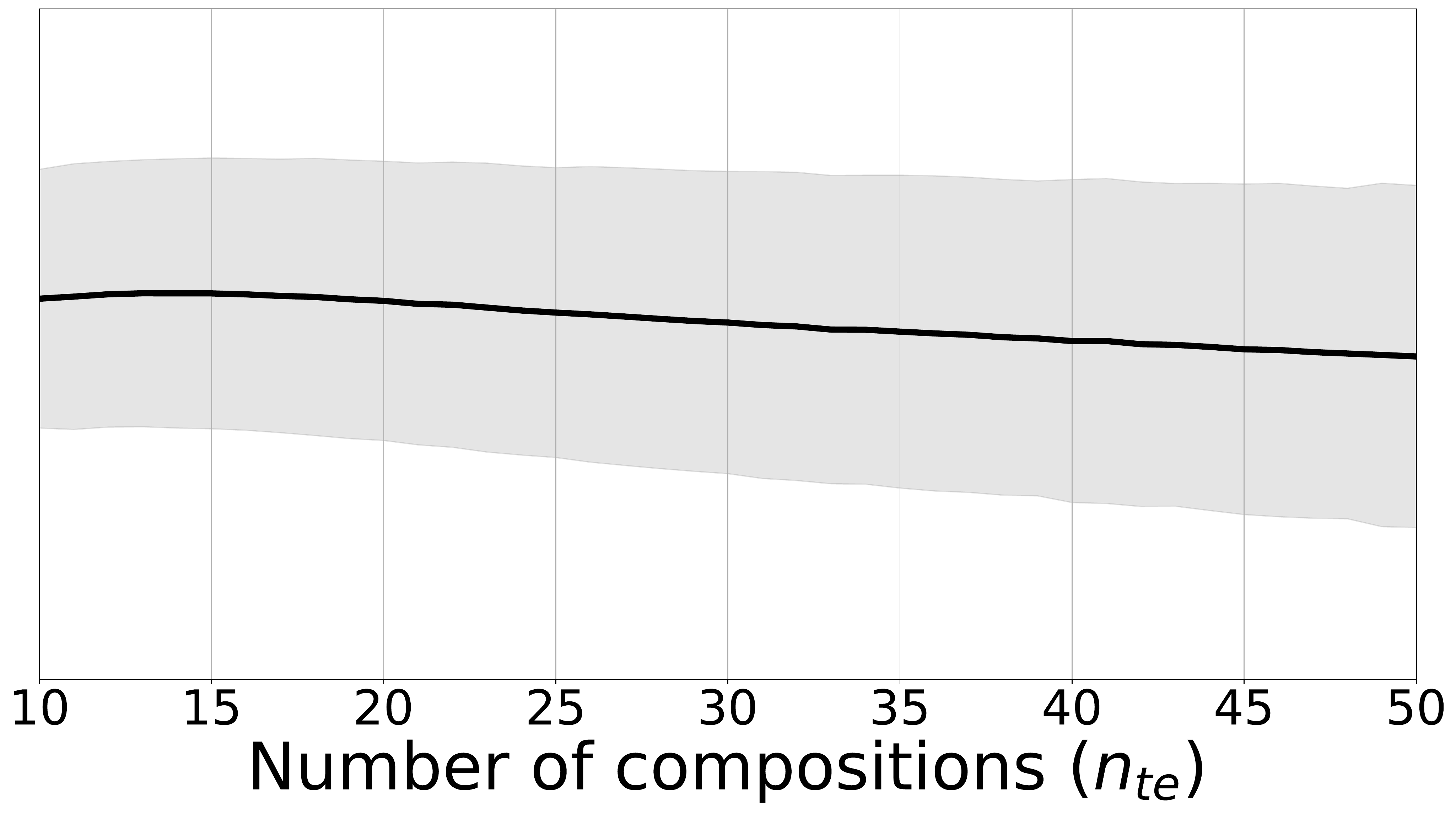}\\
\end{tabular}
}
\hfill
\begin{minipage}{0.3\linewidth}
\caption{\label{tab:ada_comp}
 Response of the discriminators as a function of the number of compositions for the transformation Gaussian noise$\rightarrow$natural image.
 We plot the average response and the standard deviation over examples (gray). 
 Higher=the discriminator classifies the image into its target domain.
\emph{Top:} training with a fixed number of compositions, $\maxcomp$=30.
\emph{Bottom:} training with randomised $\maxcomp \in [\![20, 30]\!]$. 
}
\end{minipage}
\vspace{-0.6em}
\end{figure*}

\paragraph{Analysis of the progressive training strategy.}

Table~\ref{tab:psnr_urban100_sharing_max_rep} also evaluates the impact of the maximum number of compositions $\maxcomp$.
Having several compositions clearly helps.
Since we choose the number of compositions $\maxcomptest$ at inference time (see next paragraph), it may be relevant to vary $\maxcomp$ at training time to minimize the discrepancy between the train- and test-time settings.

For this, we tried different intervals to randomly draw the maximum number of compositions for each epoch, after the warm-up phase.
If $\maxcomptest$ is fixed, the optimal choice is $\maxcomp=\maxcomptest$. However, if we use an adaptive $\maxcomptest$, the best range is $\maxcomp\in[\![20,30]\!]$, and the adaptive case with randomised training gives the best performance for denoising and debluring. 
Appendix~\ref{app:training} reports results obtained with different $\maxcomp$ ranges. 

\paragraph{Stopping criterion.} 

We consider two cases: either we use a fixed $\maxcomptest$, or we 
use the discriminator to evaluate the  transformation quality: it selects the value $\maxcomptest$ maximizing the target discriminator error for a given image.
Figure~\ref{tab:ada_comp} shows that setting a fixed $\maxcomp$ causes the discriminator to select $\maxcomptest = \maxcomp$ as the best iteration at inference time.
By selecting the best $\maxcomptest$ for each image we obtain on average 
a PSNR improvement of +1.36dB   
for a Gaussian noise of standard deviation $30$, compared to fixing $\maxcomptest$.
In Appendix~\ref{app:inference}, we compare it with the best possible stopping criterion: an Oracle that selects $\maxcomptest$ directly on the PSNR. 
Our adaptive strategy significantly tightens the gap to this  upper bound. 

\paragraph{Comparison with CycleGAN.}

We use CycleGAN as a baseline. 
The differences between CycleGAN and powers of layers are 
(1) we use a single encoder and decoder trained in advance and common to all tasks;
(2) CycleGAN has 9 residual blocks, PoL iterates a single residual block an arbitrary number of times.
The inference time of PoL depends on the number of compositions $\maxcomptest$ but the number of parameters does not:

\begin{center}
{
\vspace{-10pt}
\scalebox{0.8}{
\begin{tabular}{c|cc|c|c}
    \toprule
    & encoder + decoder & residual block & discriminators & total \\
    PoL& $1\times$ 1.7M & $2\times\expfactor \times 1.1$M & $2\times$2.7M & 15.9M\\
    CycleGAN & \multicolumn{2}{c|}{$2\times$ 11.4M}& $2\times$2.7M & 28.2M \\
    \bottomrule
\end{tabular}
}
}
\end{center}

\begin{figure*}
\centering
\scalebox{0.625}{
\small
\def \mysp {\hspace{5pt}}
\begin{tabular}{|c|ccc|c|ccc|c|ccc||c|cc|cc|cc|}
  \toprule
    \multicolumn{12}{|c||}{Impact of the noise intensity applied to Urban-100 images} & \multicolumn{7}{|c|}{Impact of the amount of training data } \\
    \midrule
    \multicolumn{8}{|c|}{PSNR (higher=better)} & \multicolumn{4}{|c||}{NIQE (lower=better)} & &\multicolumn{4}{|c|}{PSNR} & \multicolumn{2}{|c|}{NIQE}\\
    \midrule
     \multicolumn{1}{|c}{Noise} &  Noisy & \multicolumn{2}{c|}{Denoising}  &
     \multicolumn{1}{|c}{Sigma} & Blur & \multicolumn{2 }{c|}{Debluring}& 
     \multicolumn{1}{|c}{JPEG} & JPEG &\multicolumn{2}{c||}{Deblocking} &
     \# train  & \multicolumn{2}{|c|}{Denoise (std=30)} &
     \multicolumn{2}{c}{Deblur ($\sigma$=4)}& 
     \multicolumn{2}{|c|}{JPEG (qual 25)}\\

    \multicolumn{1}{|c}{(std)} &   images & CLG & PoL &
    \multicolumn{1}{|c}{Blur} &  images & CLG & PoL& 
    \multicolumn{1}{|c}{quality} &  images & CLG & PoL &
    images & CLG  & PoL & CLG  & PoL & CLG  & PoL \\
    \midrule	
    15 & 24.9 & 22.4  &\textbf{27.4}  &
    2 & 21.6 & 20.4 & \textbf{22.1}&
    15 & 9.0  & \textbf{7.9}  &\textbf{7.9} &
    1 & 14.5& \textbf{22.3}& 14.4& \textbf{18.7} & 15.6& \textbf{7.9}\\

    30 & 19.2   & 21.9 & \textbf{23.7}  & 
    4 & 19.2  & 18.5 & \textbf{19.2} &
    25 &  8.9 & 7.5  & \textbf{7.1} &
    5 & 16.6 & \textbf{23.1} & 16.7 & \textbf{18.8} & 12.6 & \textbf{7.6} \\

    50 & 15.2  & 21.6 & \textbf{ 22.5}  &
    8 &  17.5  & 16.0 & \textbf{17.5} &
    30 & 8.9  & 7.5 & \textbf{6.8}&
    10 & 20.9 & \textbf{23.2} & 17.0 & \textbf{18.8} & 7.8 & \textbf{7.3}\\

    70 & 12.7  & \textbf{21.0}& 20.9 &
    16 & 16.1  & 16.1 & \textbf{16.2}  &
    50 & 9.0  & 7.0  &   \textbf{6.6} &
    100 & 21.0 & \textbf{23.4} & 18.2 & \textbf{19.0} & 7.5& \textbf{7.0} \\

    100 & 10.4  & \textbf{20.0} & 19.4 & 
    24 & 15.5  & 12.9 & \textbf{15.5}&
    70 & 8.9 & 7.1  &   \textbf{6.8} & 
    400 & 21.8 & \textbf{23.6} & 18.2 & \textbf{19.0}& 7.5 & \textbf{6.9}\\
\bottomrule
\end{tabular}
}
\scalebox{0.65}{
    \small
    \begin{tabular}{@{}c@{\hspace{5pt}}c|c@{\hspace{5pt}}c|c@{\hspace{5pt}}c@{}}
      \multicolumn{2}{c}{Deblocking} & \multicolumn{2}{c}{Denoising} & \multicolumn{2}{c}{Debluring}    \\
     Original Image  & Image JPEG (quality=25)  & Image with Noise(std=30)  & CycleGAN & Image with Blur ($\sigma$=4) & CycleGAN\\ 
    \includegraphics[width=0.24\linewidth]{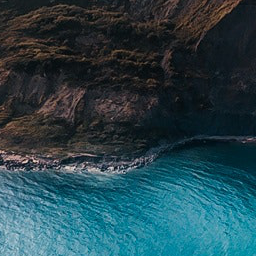} &
    \includegraphics[width=0.24\linewidth]{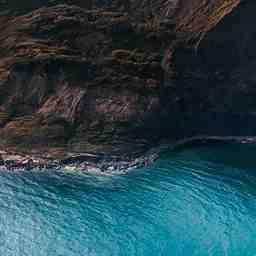} &
    
     \includegraphics[width=0.24\linewidth]{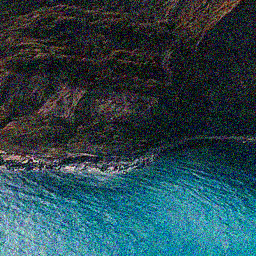} & 
     \includegraphics[width=0.24\linewidth]{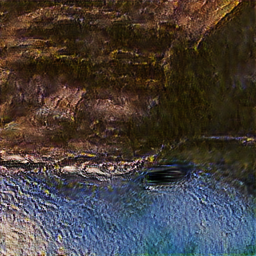}&

     \includegraphics[width=0.24\linewidth]{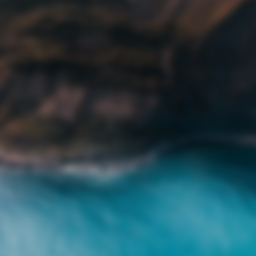}&
      \includegraphics[width=0.24\linewidth]{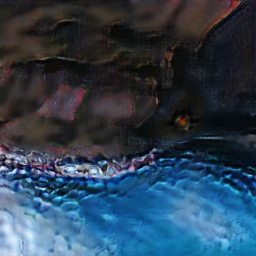}\\

      CycleGAN & PoL  & PoL (manual) & PoL (discriminator) &  PoL (manual)  & PoL (discriminator)\\
      \includegraphics[width=0.24\linewidth]{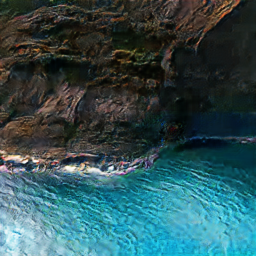}&
     \includegraphics[width=0.24\linewidth]{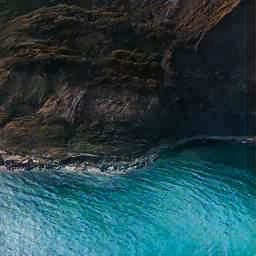} &
     
     \includegraphics[width=0.24\linewidth]{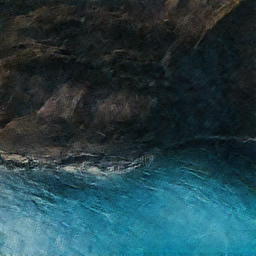}&
    \includegraphics[width=0.24\linewidth]{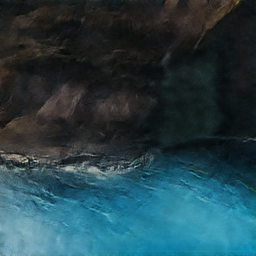} &
    
     \includegraphics[width=0.24\linewidth]{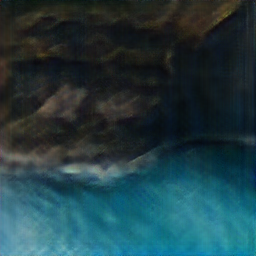}&
        \includegraphics[width=0.24\linewidth]{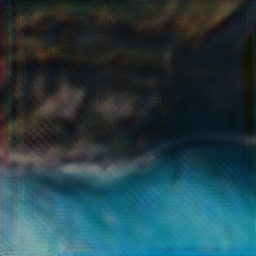} \\
    \end{tabular}

}
\caption{\label{tab:psnr_urban100_comp}
{\it Top:} Comparison between Powers of layers (PoL) and CycleGAN (CLG) to denoise images of  Urban-100~\cite{Huang2015Urban100}.
We provide standard deviation and additional results in Appendix~\ref{app:cyclegan}.
{\it Bottom:} Visual comparison between our method (manual and discriminator choice) and CycleGAN for deblocking, denoising and debluring.  
}
\end{figure*}

Figure~\ref{tab:psnr_urban100_comp} compares the performance obtained by the two methods on denoising tasks with varying noise intensities. 
PoL gives better results than CycleGAN in terms of objective metrics, and overall the  images produced by our method look as realistic and/or accurate.

\paragraph{Training with few images.}
 
Figure~\ref{tab:psnr_urban100_comp} also compares our method with CycleGAN  when training in a data-starving scenario. 
Whatever the number of  training images, PoL outperforms CycleGAN, but the gap is particularly important with very few images. 
This is expected for two reasons. Firstly, our approach has fewer parameters to learn than CycleGAN. Secondly, it only requires to learn the transformation because the encoder and decoder are pre-learned as a vanilla auto-encoder, while CycleGAN needs to learn how to encode and decode images.

\paragraph{Parametrization: remarks.}

Beyond the settings inherited from CycleGAN, the main training parameters of Powers of layers are the maximum number of compositions $\maxcomp$ and the range from which they are randomly sampled.
The number of compositions at inference time $\maxcomptest$ is also important but the discriminator criterion can be used to set it automatically.

\section{Experiments}\label{sec:experiments}

We now run experiments on two image generation applications. We refer to Section~\ref{sec:arch} for the architecture and training protocol. 

In Appendix~\ref{app:cyclegan} we also give a comparison with the Fader network for the capacity to modulate a transformation, and more visual examples in Appendix~\ref{app:visu}. 

\begin{figure*}

    \centering
    \scalebox{0.8}{
    \small
    \begin{tabular}{lcc}
      \toprule
      Domain &  CycleGAN  &   POL\\
    \midrule	
    Summer $\rightarrow$ Winter & 48.8 & \textbf{46.1} \\
    Summer $\leftarrow$ Winter & 48.4  & \textbf{44.4} \\
    \midrule	
    Horse $\rightarrow$ Zebra & 89.7  & \textbf{53.0} \\
    Horse $\leftarrow$ Zebra & \textbf{110.5} & 112.3 \\
    \midrule	
    Van-Gogh $\rightarrow$ Picture  & 163.4  & \textbf{134.4} \\
    Van-Gogh $\leftarrow$ Picture  & \textbf{151.4} & 152.7 \\
    \midrule	
    Cezanne $\rightarrow$ Picture &\textbf{127.4}  & 138.8 \\
    Cezanne $\leftarrow$ Picture & \textbf{145.5}  & 147.6 \\
    \midrule	
    Monet $\rightarrow$ Picture & \textbf{60.3}  & 70.3 \\
    Monet $\leftarrow$ Picture & \textbf{61.8}  & 82.1 \\
    \midrule	
    Apple $\rightarrow$ Orange & 88.9  & \textbf{83.2} \\
    Apple $\leftarrow$ Orange & 116.7  & \textbf{113.2} \\
    \bottomrule
    \end{tabular}\hspace{1cm}%
      \scalebox{0.96}{
  \small
  \newcommand{\cycleganim}[1]{\includegraphics[width=0.15\linewidth]{images/CyCleGAN_Visual/#1}}
  \begin{tabular}{ccccc}
  \toprule
&  horse$\rightarrow$zebra  & 
 summer$\rightarrow$winter &
 Monet$\rightarrow$photo &
 orange$\rightarrow$apple\\ 
    \rotatebox{90}{Original image} &
    \cycleganim{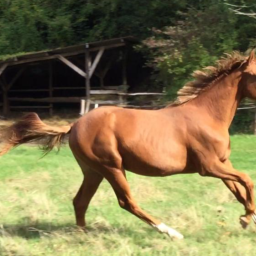} &
    \cycleganim{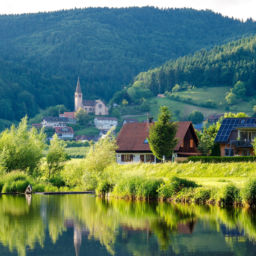} &
    \cycleganim{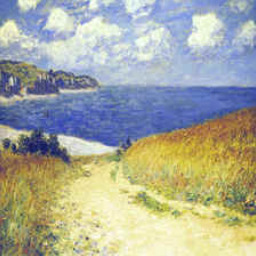} &
    \cycleganim{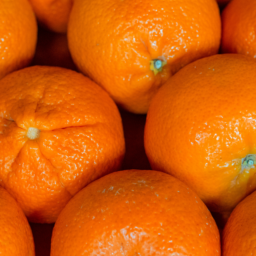}\\

    \midrule

    \rotatebox{90}{CycleGAN} &
    \cycleganim{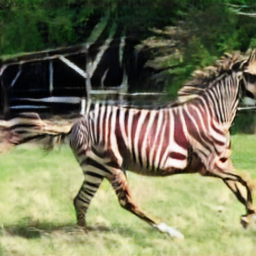} &
    \cycleganim{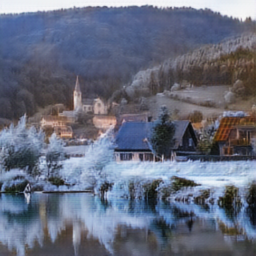}&
    \cycleganim{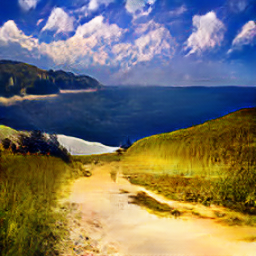}&
    \cycleganim{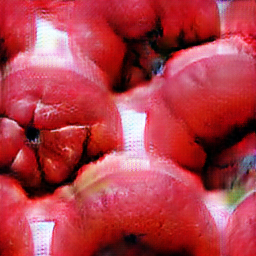}\\

    \midrule
    \rotatebox{90}{Powers of layers} &
    \cycleganim{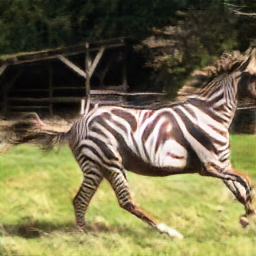} &
    \cycleganim{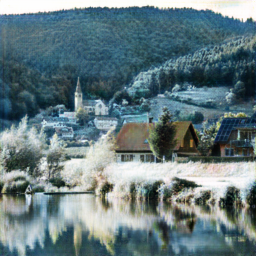}&
    \cycleganim{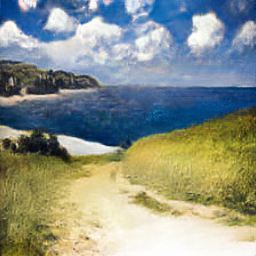}&
    \cycleganim{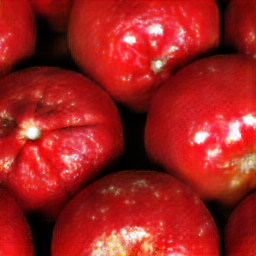}\\
  \end{tabular}
  }

    }
    \caption{\label{tab:cycleganfid}
     Left: Frechet Inception Distance (FID) obtained on image-to-image translation tasks (lower is better).
     We compare CycleGAN and our Powers of layers method.
     Right: example results.
    }
    \end{figure*}

\paragraph{Unpaired image-to-image translation.}
We report results for 6 of the 8 unpaired image-to-image translation tasks introduced in the CycleGAN~\cite{Zhu2017CycleGAN} paper (the two remaining ones lead to the same conclusions) and we used the datasets from the website~\cite{cycleGANtasks}. 
We compare the Frechet Inception Distance (FID) of these two approaches in Figure~\ref{tab:cycleganfid}.
The FID measures the similarity between two datasets of images, we use it to compare the target dataset with the transformed dataset. 
It is a noisy measure for which only large deviations are significant. 
Yet the results and visualization show that our method has results comparable to those of CycleGAN, achieved with much fewer parameters.

\begin{figure*}
\centering

\newcommand{\imHR}[1]{\includegraphics[width=0.25\linewidth]{images/High_Resolution/#1}}
\newcommand{\drawat}[3]{\makebox[0pt][l]{\raisebox{#2}{\hspace*{#1}#3}}}
\newcommand{\imHRsub}[1]{\drawat{-35mm}{0mm}{%
\fcolorbox{white}{white}{\includegraphics[width=0.1\linewidth]{images/High_Resolution/#1}}
}}

\scalebox{0.86}{
\small
\begin{tabular}{@{}ccc@{\hspace{0.7cm}}|}
 Photo $\rightarrow$ Van Gogh & Photo $\rightarrow$ Monet & Horse $\rightarrow$ Zebra \\
 \imHR{VanGogh/im_apple_change_Human_crop_12.jpg}%
 \imHRsub{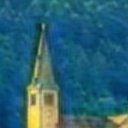} &
 \imHR{Monet/im_monet_change_Human_crop_20.jpg}
   &\imHR{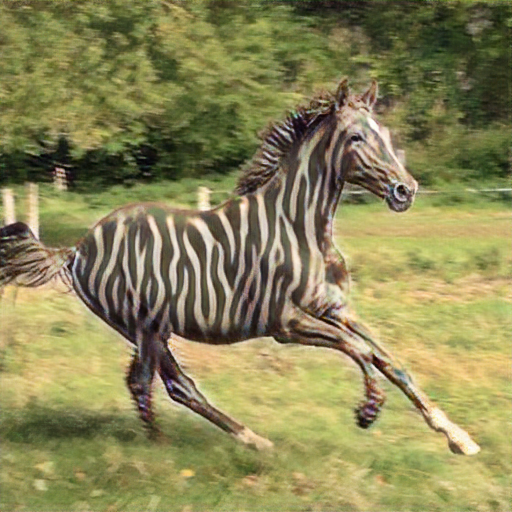}\\
  \imHR{VanGogh/im_apple_change_Human_crop_16.jpg}
  &\imHR{Monet/im_monet_change_Human_crop_18.jpg} 
  &\imHR{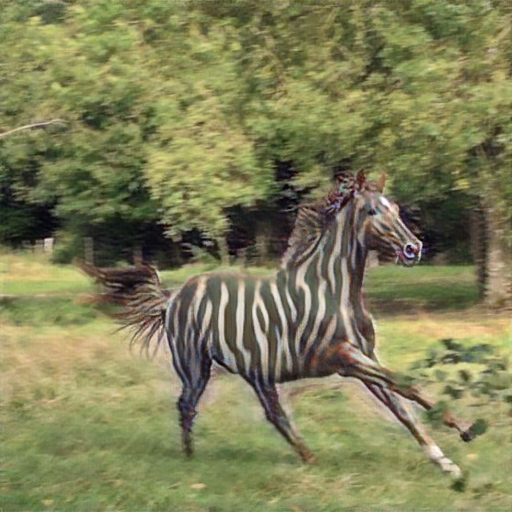}\\
\end{tabular}
\hspace{0.4cm}%
    \centering
\small
\scalebox{0.8}{
\begin{tabular}{c}
Trained at $512\times 512$  \\
 \includegraphics[width=0.3\linewidth]{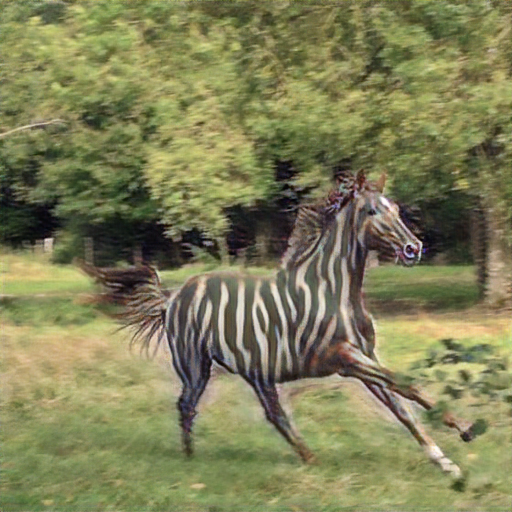}\\
  Trained at $256\times 256$\\
 \includegraphics[width=0.3\linewidth]{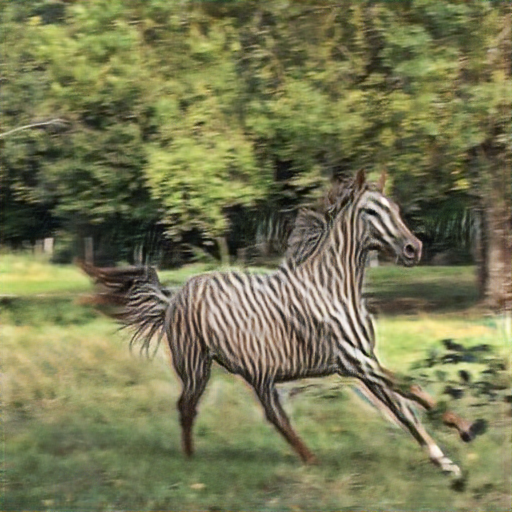}\\
\end{tabular}
}

}
\caption{\label{fig:highres}
{\em Left:} Different visual results with high resolution images.
See the original images in Appendix \ref{app:visu}.
{\em Right:} Comparisons of the generations obtained by models trained either with high-resolution or low-resolution images, applied to a high-resolution test image.
}
\end{figure*}

\paragraph{High resolution experiments.}

PoL is fully convolutional architecture, therefore it is technically possible to apply models trained in a low resolution to high resolution images.
However, the results are not always convincing, as shown in  Figure~\ref{fig:highres} (right) where the model trained on low resolution images does not create stripes at the ``right'' scale on zebras.
To circumvent this problem, CycleGAN trains on patches taken from high resolution images.
This works for transformations affecting the whole image (painting$\leftrightarrow$photo), but this is not applicable in the case where only a part of the image is affected (horse$\rightarrow$zebra).
In contrast, our proposal can adapt the memory used by changing its number of compositions, so we can apply it to very large images without running out of memory.
Figure~\ref{fig:highres} (left) depicts results obtained with our method trained on high resolution images.

\paragraph{Combining transformation.} 

\begin{figure*}
\vspace{-0.2em}
\centering
\scalebox{1}{
\small
\begin{tabular}{@{}c@{\hspace{20pt}}|@{\hspace{20pt}}c@{\hspace{10pt}}c@{\hspace{20pt}}|@{\hspace{20pt}}c@{\hspace{10pt}}c@{}}
  \multicolumn{3}{c}{\hspace{50pt} Composition in embedding space}  &\multicolumn{2}{c}{Composition in image space} \\
 Original & +denoising & +to zebra & +denoising & +to zebra \\
\includegraphics[width=0.125\linewidth]{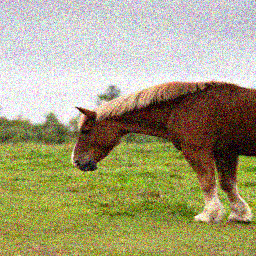}&
\includegraphics[width=0.125\linewidth]{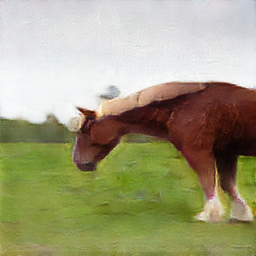}
 &\includegraphics[width=0.125\linewidth]{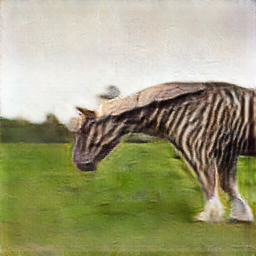}&
 
 \includegraphics[width=0.125\linewidth]{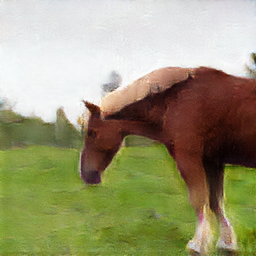}
 &\includegraphics[width=0.125\linewidth]{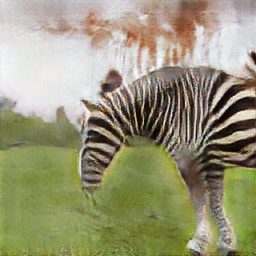}\\
 
    \includegraphics[width=0.125\linewidth]{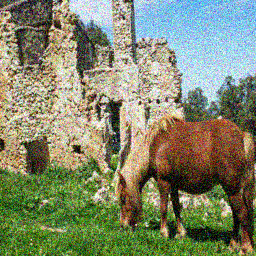}&
 \includegraphics[width=0.125\linewidth]{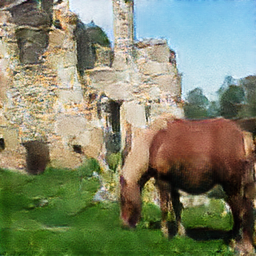}
   &\includegraphics[width=0.125\linewidth]{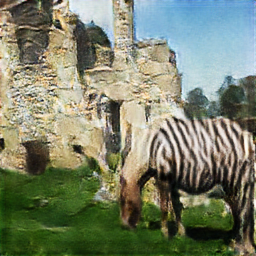}&
   
    \includegraphics[width=0.125\linewidth]{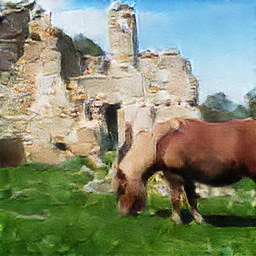}
   &\includegraphics[width=0.125\linewidth]{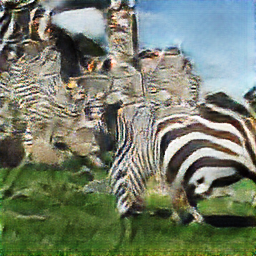}\\
\end{tabular}
}
\vspace{-5pt}
\caption{\label{fig:composition_embedding_space}
Composition of transformations in the embedding/image space.
}
\vspace{-0.6em}
\end{figure*}

The different blocks associated with different transformations operate in the same embedding space for different tasks. Hence we can compose transformations, each being realized by one residual block.
We train Transform~\#1 in the usual way, then freeze its residual block.
Transform~\#2 is trained on the output of \#1. 
Visual results are in Figure~\ref{fig:composition_embedding_space}.
The composition in the embedding space gives better results than decoding/encoding to image space mid-way. 

\section{Conclusion}

Powers of layers consists in iterating a residual block to learns a complex transformation with no direct supervision. 
On various tasks, power of layers gives similar performance to CycleGAN with fewer parameters. 
The flexibility offered by the common embedding space can be used to modulate the strength of a transformation or to compose several transformations.
While in most examples the discriminator is only used for training, 
Powers of layers can also exploit it to adjust the transformation to the input image at inference time.

\clearpage
\bibliographystyle{plain}
\bibliography{egbib}

\begin{thebibliography}{10}

\bibitem{Agustsson_2017_CVPR_Workshops}
Eirikur Agustsson and Radu Timofte.
\newblock Ntire 2017 challenge on single image super-resolution: Dataset and
  study.
\newblock In {\em The IEEE Conference on Computer Vision and Pattern
  Recognition (CVPR) Workshops}, 2017.

\bibitem{barnsley1988science}
Michael~F Barnsley, Robert~L Devaney, Benoit~B Mandelbrot, Heinz-Otto Peitgen,
  Dietmar Saupe, Richard~F Voss, Yuval Fisher, and Michael McGuire.
\newblock {\em The science of fractal images}.
\newblock Springer, 1988.

\bibitem{Bousmalis2016DomainSN}
Konstantinos Bousmalis, George Trigeorgis, Nathan Silberman, Dilip Krishnan,
  and Dumitru Erhan.
\newblock Domain separation networks.
\newblock In {\em Advances in Neural Information Processing Systems}, 2016.

\bibitem{Brock2018LargeSG}
Andrew Brock, Jeff Donahue, and Karen Simonyan.
\newblock Large scale gan training for high fidelity natural image synthesis.
\newblock {\em International Conference on Learning Representations}, 2018.

\bibitem{chen2017fast}
Qifeng Chen, Jia Xu, and Vladlen Koltun.
\newblock Fast image processing with fully-convolutional networks.
\newblock In {\em Proceedings of the IEEE International Conference on Computer
  Vision}, pages 2497--2506, 2017.

\bibitem{Chen2018NeuralOD}
Tian~Qi Chen, Yulia Rubanova, Jesse Bettencourt, and David Duvenaud.
\newblock Neural ordinary differential equations.
\newblock In {\em Advances in Neural Information Processing Systems}, 2018.

\bibitem{choi2019starganv2}
Yunjey Choi, Youngjung Uh, Jaejun Yoo, and Jung-Woo Ha.
\newblock Stargan v2: Diverse image synthesis for multiple domains.
\newblock {\em Conference on Computer Vision and Pattern Recognition}, 2020.

\bibitem{conneau2017word}
Alexis Conneau, Guillaume Lample, Marc'Aurelio Ranzato, Ludovic Denoyer, and
  Herv{\'e} J{\'e}gou.
\newblock Word translation without parallel data.
\newblock {\em arXiv preprint arXiv:1710.04087}, 2017.

\bibitem{Figurnov2016SpatiallyAC}
Michael Figurnov, Maxwell~D. Collins, Yukun Zhu, Li~Zhang, Jonathan Huang,
  Dmitry~P. Vetrov, and Ruslan Salakhutdinov.
\newblock Spatially adaptive computation time for residual networks.
\newblock {\em Conference on Computer Vision and Pattern Recognition}, 2017.

\bibitem{Fu2019GeometryConsistentGA}
Huan Fu, Mingming Gong, Chaohui Wang, Kayhan Batmanghelich, Kun Zhang, and
  Dacheng Tao.
\newblock Geometry-consistent generative adversarial networks for one-sided
  unsupervised domain mapping.
\newblock {\em Conference on Computer Vision and Pattern Recognition}, 2019.

\bibitem{Gao2019CrossDM}
Shangqian Gao, Cheng Deng, and Heng Huang.
\newblock Cross domain model compression by structurally weight sharing.
\newblock {\em Conference on Computer Vision and Pattern Recognition}, 2019.

\bibitem{goodfellow2016deep}
Ian Goodfellow, Yoshua Bengio, and Aaron Courville.
\newblock {\em Deep learning}.
\newblock MIT press, 2016.

\bibitem{Goodfellow2014GenerativeAN}
Ian~J. Goodfellow, Jean Pouget-Abadie, Mehdi Mirza, Bing Xu, David
  Warde-Farley, Sherjil Ozair, Aaron~C. Courville, and Yoshua Bengio.
\newblock Generative adversarial nets.
\newblock In {\em Advances in Neural Information Processing Systems}, 2014.

\bibitem{Han2015DeepCC}
Song Han, Huizi Mao, and William~J. Dally.
\newblock Deep compression: Compressing deep neural network with pruning,
  trained quantization and huffman coding.
\newblock {\em CoRR}, 2015.

\bibitem{hornik1989multilayer}
Kurt Hornik, Maxwell Stinchcombe, Halbert White, et~al.
\newblock Multilayer feedforward networks are universal approximators.
\newblock {\em Neural networks}, 2(5):359--366, 1989.

\bibitem{Huang2016DeepNW}
Gao Huang, Yu~Sun, Zhuang Liu, Daniel Sedra, and Kilian~Q. Weinberger.
\newblock Deep networks with stochastic depth.
\newblock In {\em European Conference on Computer Vision}, 2016.

\bibitem{Huang2015Urban100}
Jia-Bin Huang, Abhishek Singh, and Narendra Ahuja.
\newblock Single image super-resolution from transformed self-exemplars.
\newblock In {\em Conference on Computer Vision and Pattern Recognition}, 2015.

\bibitem{huang2018munit}
Xun Huang, Ming-Yu Liu, Serge Belongie, and Jan Kautz.
\newblock Multimodal unsupervised image-to-image translation.
\newblock In {\em European Conference on Computer Vision}, 2018.

\bibitem{Jastrzebski2017ResidualCE}
Stanislaw Jastrzebski, Devansh Arpit, Nicolas Ballas, Vikas Verma, Tong Che,
  and Yoshua Bengio.
\newblock Residual connections encourage iterative inference.
\newblock {\em International Conference on Learning Representations}, 2017.

\bibitem{Jeon2020DifferentiableFI}
Younahan Jeon, Minsik Lee, and Jin~Young Choi.
\newblock Differentiable fixed-point iteration layer.
\newblock {\em arXiv preprint arXiv:2002.02868}, 2020.

\bibitem{Karnewar2019MSGGANMG}
Animesh Karnewar, Oliver Wang, and Raghu~Sesha Iyengar.
\newblock Msg-gan: Multi-scale gradient gan for stable image synthesis.
\newblock {\em arXiv preprint arXiv:1903.06048}, 2019.

\bibitem{karras2017progressive}
Tero Karras, Timo Aila, Samuli Laine, and Jaakko Lehtinen.
\newblock Progressive growing of gans for improved quality, stability, and
  variation.
\newblock In {\em International Conference on Learning Representations}, 2017.

\bibitem{Karras2018ASG}
Tero Karras, Samuli Laine, and Timo Aila.
\newblock A style-based generator architecture for generative adversarial
  networks.
\newblock {\em Conference on Computer Vision and Pattern Recognition}, 2019.

\bibitem{Karras2019AnalyzingAI}
Tero Karras, Samuli Laine, Miika Aittala, Janne Hellsten, Jaakko Lehtinen, and
  Timo Aila.
\newblock Analyzing and improving the image quality of stylegan.
\newblock {\em arXiv preprint arXiv:1912.04958}, 2019.

\bibitem{lample2017fader}
Guillaume Lample, Neil Zeghidour, Nicolas Usunier, Antoine Bordes, Ludovic
  Denoyer, et~al.
\newblock Fader networks: Manipulating images by sliding attributes.
\newblock In {\em Advances in Neural Information Processing Systems}, 2017.

\bibitem{Lin2019COCOGANGB}
Chieh~Hubert Lin, Chia-Che Chang, Yu-Sheng Chen, Da-Cheng Juan, Wei Wei, and
  Hwann-Tzong Chen.
\newblock Coco-gan: Generation by parts via conditional coordinating.
\newblock {\em International Conference on Computer Vision}, 2019.

\bibitem{Liu2017UnsupervisedIT}
Ming-Yu Liu, Thomas Breuel, and Jan Kautz.
\newblock Unsupervised image-to-image translation networks.
\newblock In {\em Advances in Neural Information Processing Systems}, 2017.

\bibitem{liu2019FUNIT}
Ming-Yu Liu, Xun Huang, Arun Mallya, Tero Karras, Timo Aila, Jaakko Lehtinen,
  and Jan Kautz.
\newblock Few-shot unsupervised image-to-image translation.
\newblock In {\em International Conference on Computer Vision}, 2019.

\bibitem{Liu2016CoupledGA}
Ming-Yu Liu and Oncel Tuzel.
\newblock Coupled generative adversarial networks.
\newblock In {\em Advances in Neural Information Processing Systems}, 2016.

\bibitem{Mittal2013NIQE}
Anish Mittal, Rajiv Soundararajan, and Alan~C. Bovik.
\newblock Making a “completely blind” image quality analyzer.
\newblock {\em IEEE Signal Processing Letters}, 2013.

\bibitem{montufar2014universal}
Guido~F Mont{\'u}far.
\newblock Universal approximation depth and errors of narrow belief networks
  with discrete units.
\newblock {\em Neural computation}, 26(7):1386--1407, 2014.

\bibitem{park2019SPADE}
Taesung Park, Ming-Yu Liu, Ting-Chun Wang, and Jun-Yan Zhu.
\newblock Semantic image synthesis with spatially-adaptive normalization.
\newblock In {\em Conference on Computer Vision and Pattern Recognition}, 2019.

\bibitem{Pathak2016ContextEF}
Deepak Pathak, Philipp Kr{\"a}henb{\"u}hl, Jeff Donahue, Trevor Darrell, and
  Alexei~A. Efros.
\newblock Context encoders: Feature learning by inpainting.
\newblock {\em Conference on Computer Vision and Pattern Recognition}, 2016.

\bibitem{Polino2018ModelCV}
Antonio Polino, Razvan Pascanu, and Dan Alistarh.
\newblock Model compression via distillation and quantization.
\newblock {\em International Conference on Learning Representations}, 2018.

\bibitem{Radford2015UnsupervisedRL}
Alec Radford, Luke Metz, and Soumith Chintala.
\newblock Unsupervised representation learning with deep convolutional
  generative adversarial networks.
\newblock {\em International Conference on Learning Representations}, 2015.

\bibitem{rottshaham2019singan}
Tamar Rott~Shaham, Tali Dekel, and Tomer Michaeli.
\newblock Singan: Learning a generative model from a single natural image.
\newblock In {\em International Conference on Computer Vision}, 2019.

\bibitem{Thomee2016YFCC100MTN}
Bart Thomee, David~A. Shamma, Gerald Friedland, Benjamin Elizalde, Karl Ni,
  Douglas Poland, Damian Borth, and Lijia Li.
\newblock Yfcc100m: the new data in multimedia research.
\newblock {\em Commun. ACM}, 2016.

\bibitem{Vaswani2017AttentionIA}
Ashish Vaswani, Noam Shazeer, Niki Parmar, Jakob Uszkoreit, Llion Jones,
  Aidan~N. Gomez, Lukasz Kaiser, and Illia Polosukhin.
\newblock Attention is all you need.
\newblock In {\em Advances in Neural Information Processing Systems}, 2017.

\bibitem{Veit2017ConvolutionalNW}
Andreas Veit and Serge~J. Belongie.
\newblock Convolutional networks with adaptive inference graphs.
\newblock In {\em European Conference on Computer Vision}, 2017.

\bibitem{viazovetskyi2020stylegan2}
Yuri Viazovetskyi, Vladimir Ivashkin, and Evgeny Kashin.
\newblock Stylegan2 distillation for feed-forward image manipulation, 2020.

\bibitem{wang2018pix2pixHD}
Ting-Chun Wang, Ming-Yu Liu, Jun-Yan Zhu, Andrew Tao, Jan Kautz, and Bryan
  Catanzaro.
\newblock High-resolution image synthesis and semantic manipulation with
  conditional gans.
\newblock In {\em Conference on Computer Vision and Pattern Recognition}, 2018.

\bibitem{Wu2017BlockDropDI}
Zuxuan Wu, Tushar Nagarajan, Abhishek Kumar, Steven Rennie, Larry~S. Davis,
  Kristen Grauman, and Rog{\'e}rio~Schmidt Feris.
\newblock Blockdrop: Dynamic inference paths in residual networks.
\newblock {\em Conference on Computer Vision and Pattern Recognition}, 2017.

\bibitem{Yi2017DualGAN}
Zili Yi, Hao Zhang, Ping Tan, and Minglun Gong.
\newblock Dualgan: Unsupervised dual learning for image-to-image translation.
\newblock {\em International Conference on Computer Vision}, pages 2868--2876,
  2017.

\bibitem{Zakharov2019FewShotAL}
Egor Zakharov, Aliaksandra Shysheya, Egor Burkov, and Victor~S. Lempitsky.
\newblock Few-shot adversarial learning of realistic neural talking head
  models.
\newblock {\em arXiv preprint arXiv:1905.08233}, 2019.

\bibitem{Zbontar2018fastMRIAO}
Jure Zbontar, Florian Knoll, Anuroop Sriram, Matthew~J. Muckley, Mary Bruno,
  Aaron Defazio, Marc Parente, Krzysztof~J Geras, Joe Katsnelson, Hersh
  Chandarana, Zizhao Zhang, Michal Drozdzal, Adriana Romero, Michael~G. Rabbat,
  Pascal Vincent, James Pinkerton, Duo Wang, Nafissa Yakubova, Erich Owens,
  C.~Lawrence Zitnick, Michael~P. Recht, Daniel~K. Sodickson, and Yvonne~W.
  Lui.
\newblock fastmri: An open dataset and benchmarks for accelerated mri.
\newblock {\em ArXiv preprint arXiv:1811.08839}, 2018.

\bibitem{Zhang2018RecurrentCA}
Zhendong Zhang and Cheolkon Jung.
\newblock Recurrent convolutions: A model compression point of view.
\newblock {\em NIPS Workshops: Compact Deep Neural Network Representation with
  Industrial Applications}, 2018.

\bibitem{cycleGANtasks}
Jun-Yan Zhu, Taesung Park, Phillip Isola, and Alexei~A Efros.
\newblock Cyclegan tasks.
\newblock
  \url{https://people.eecs.berkeley.edu/~taesung\_park/CycleGAN/datasets/}.
\newblock Accessed: 2019-12-20.

\bibitem{Zhu2017CycleGAN}
Jun-Yan Zhu, Taesung Park, Phillip Isola, and Alexei~A Efros.
\newblock Unpaired image-to-image translation using cycle-consistent
  adversarial networks.
\newblock In {\em International Conference on Computer Vision}, 2017.

\end{thebibliography}
\clearpage \newpage \appendix\newpage

\begin{center}
{\LARGE
Supplementary material for \\
``Powers of layers for image-to-image translation'' 
\vspace{1cm} \newline 
}

\textbf{Hugo Touvron, Matthijs Douze, Matthieu Cord, Herv\'e J\'egou \vspace{1em}}
\end{center}
\vspace*{1cm}

In this supplementary material we report additional analyses, results and examples that complement our paper. 
In Appendix~\ref{app:training} we consider the training phase, which supports the importance of our progressive training strategy compared to one with a fixed number of iterations. 
Appendix~\ref{app:inference} considers the inference-time choices, in particular possible strategies to select the number of iterations. 
Appendix~\ref{app:cyclegan} provides additional comparison to CycleGAN and makes a comparison with the Fader network. 
Finally we present additional visual results for high resolution images and illustrate the progressive evaluation of results along iterations in the appendix~\ref{app:visu}. 

\section{Analysis of our progressive training strategy}
\label{app:training}

\begin{figure}[b]
\centering
\includegraphics[width=.8\linewidth]{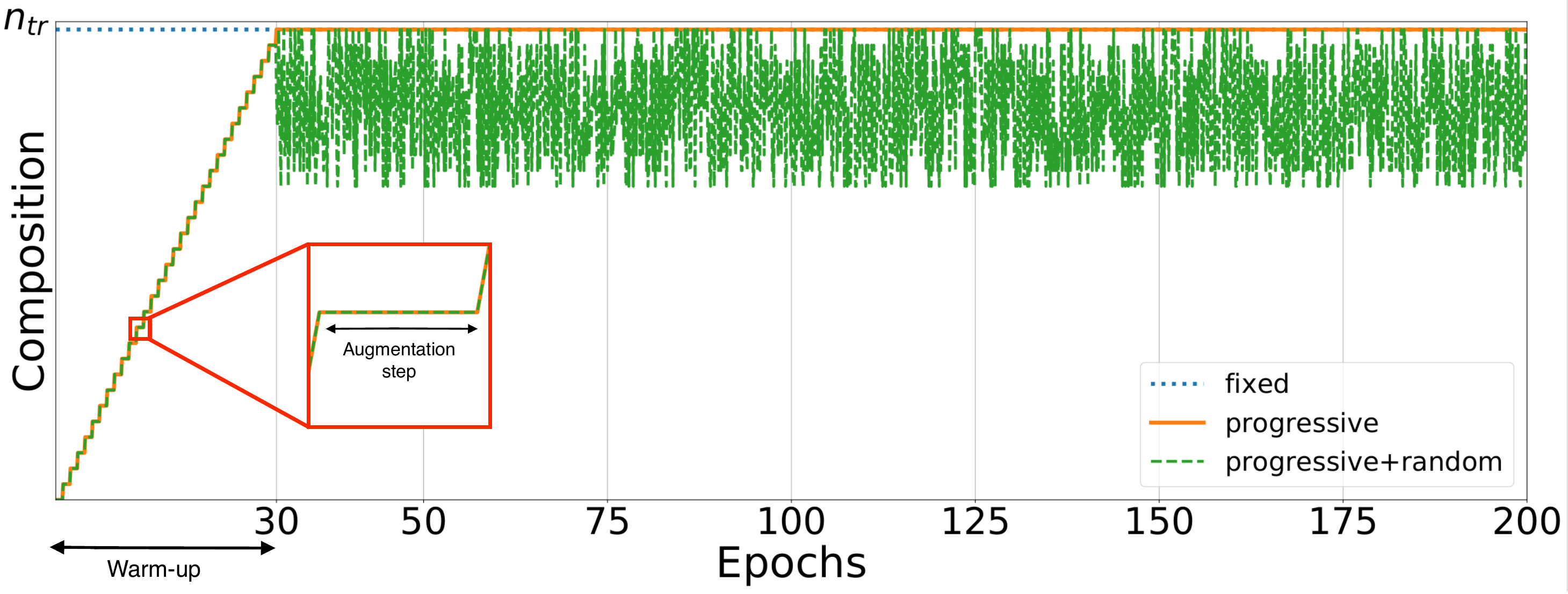}

\caption{\label{fig:illustration_training}
  Number of compositions at training time for different training strategies.
  The number of compositions is adjusted per batch. 
}
\end{figure}

Figure~\ref{fig:illustration_training} shows the different training strategies we explore. 
The degree of freedom that we can adjust is the number of compositions. 
It can be set independently per training mini-batch.

\paragraph{Progressive training versus fixed training.}

\begin{figure}[b]
\centering
\scalebox{0.85}
{
\small
\newcommand{\contHZ}[1]{\includegraphics[width=0.117\linewidth]{images/continuous/Horse_Zebra2/im_Zebra_Horse_#1.png}}
\newcommand{\noncontHZ}[1]{\includegraphics[width=0.117\linewidth]{images/not_continuous/Horse/im_not_prog_Horse_#1.png}}

\fboxrule=1pt
\fboxsep=0.1mm

\def \mysp {\hspace{2pt}}
\hspace{-20pt}
\begin{tabular}{lc@{\mysp}c@{\mysp}c@{\mysp}c@{\mysp}c@{\mysp}c@{\mysp}c@{\mysp}c@{\mysp}c@{\mysp}}
\multirow{2}{*}{
\rotatebox{90}{
\begin{minipage}[c]{1.5cm}
\centering \small Progressive \newline training
\end{minipage}
}}
& 
0& 5 & 10 &15 & 20 & 25 & 30 & 35 &40\\
& 
\contHZ{0} & 
\contHZ{5} & 
\contHZ{10} & 
\contHZ{15} & 
\contHZ{20} & 
\contHZ{25} & 
\fcolorbox{red}{red}{\contHZ{30}} & 
\contHZ{35} & 
\contHZ{40} \\
\multirow{2}{*}{
\rotatebox{90}{
\begin{minipage}[c]{1.5cm}
\centering \small fixed \newline training
\end{minipage}
}}
& 0& 5 & 10 &15 & 20 & 25 & 30 & 35 &40\\
 & 
\noncontHZ{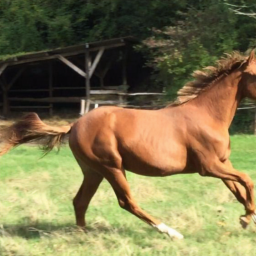} & 
\noncontHZ{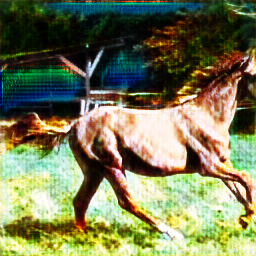} & 
\noncontHZ{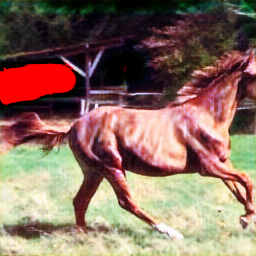} & 
\noncontHZ{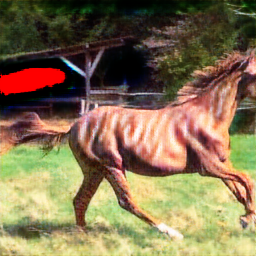} & 
\noncontHZ{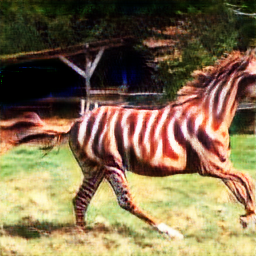} & 
\noncontHZ{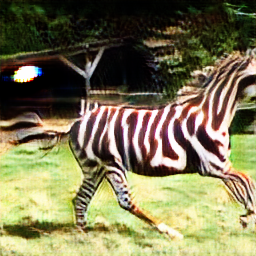} & 
\fcolorbox{red}{red}{\noncontHZ{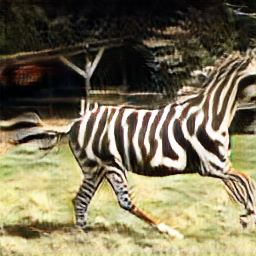}}& 
\noncontHZ{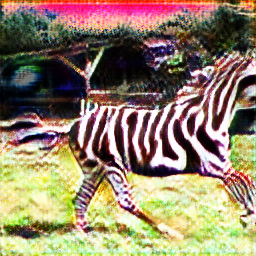} & 
\noncontHZ{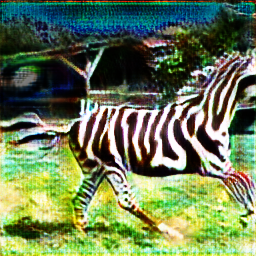} \\
\end{tabular}
}
\caption{\label{fig:schem_continuous_not_continuous}
Comparison between our progressive training approach and a non progressive approach. 
We represent the images obtained by varying the number of iterations at inference time $\maxcomptest$ in the network that transforms from domain A (horse) into domain B (zebra).
The first image ($\maxcomptest$=0) is the original image. 
Since our method was learned with $\maxcomp$=30 compositions the last two images are extrapolations. 
Our progressive training is key to ensure that all outputs look like natural images and therefore that we can modulate transformation strength at inference time.}
\end{figure}

Figure~\ref{fig:schem_continuous_not_continuous} 
illustrates the modulation of the horse$\rightarrow$zebra transformation.
This is effective only with our progressive learning, which forces the network to produce acceptable intermediate states. 

If the network is trained with a fixed number of compositions, the intermediate states do not correspond to modulations of the transformation.
The output is satisfactory only when the $\maxcomptest=\maxcomp$. In all other cases we observe artifacts in images, which therefore do not qualify as natural images. 
The generated images do not look right, and the source and target discriminators would not accept them as real images.

In contrast, with our progressive training, each number of iterations  produces a satisfactory output. Iterating the residual block gradually transforms the horse into a zebra. 

\paragraph{Warm-up phase.}
Figure~\ref{fig:PSNR_training_fix} compares the performance obtained during the first three epochs of learning with and without progressive training, with different maximum number of compositions $\maxcomp$. 
We compare this way of stabilizing the training with another classical approach: reducing the ranges of initialisation of the residual blocks. 
Figure~\ref{fig:PSNR_training_fix} shows that changing the initialization improves the performance during the first epoch, but that a warm-up phase with progressive training is more effective to improve the optimization stability. 
Figure~\ref{fig:illustration_training} shows the evolution of the number of compositions for different training strategies.

\definecolor{amber}{rgb}{1.0, 0.75, 0.0}
\definecolor{darkgreen}{rgb}{0, 0.5, 0.0}

\begin{figure*}
\begin{minipage}{0.5\linewidth}
\includegraphics[width=\linewidth]{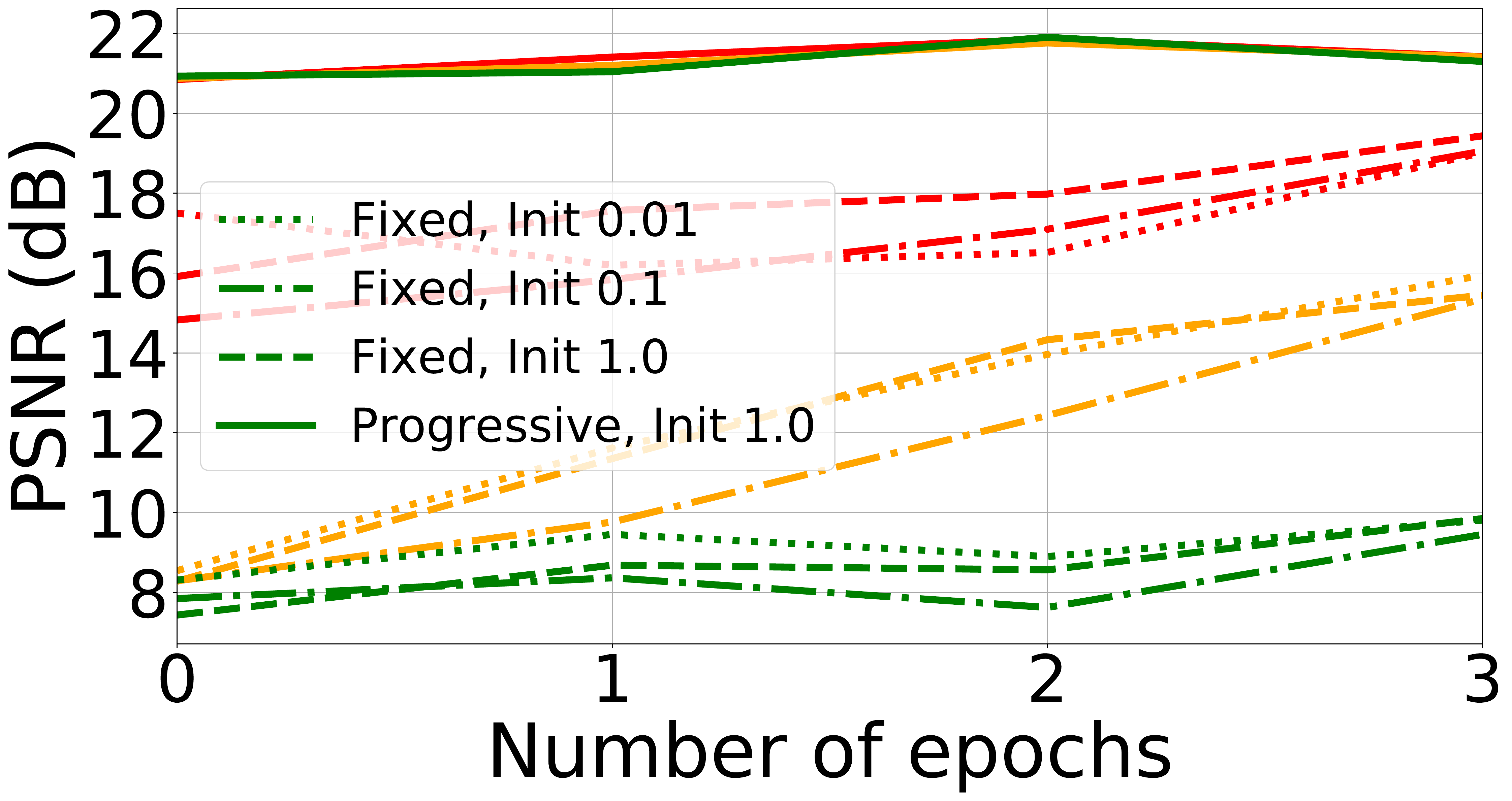}
\end{minipage}
\hfill
\begin{minipage}{0.4\linewidth}
\caption{\label{fig:PSNR_training_fix}
    Difference in PSNR between fixed learning and progressive learning during the first training epochs, evaluated on a denoising task. \newline
    \textcolor{darkgreen}{Green}: $\maxcomp$=30 \newline
    \textcolor{amber}{orange}: $\maxcomp$=16 \newline
    \textcolor{red}{red}: $\maxcomp$=4
}
\end{minipage}
\end{figure*}

\paragraph{Block composition or independent successive blocks?}
Table~\ref{tab:psnr_urban100_sharing} provides results that complement Table~\ref{tab:psnr_urban100_sharing_max_rep} in the main paper. It compares the performance with (1) the composition of the same block or (2) using independent residual blocks. In particular, we report standard deviations that assess the statistical significance of our improvement. 

\begin{table}
\caption{\label{tab:psnr_urban100_sharing}
PSNR on Urban-100. 
Comparison between our choice (PoL) and independent blocks. 
}
\vspace{0.5em}
\centering
{
\small
\begin{tabular}{|c|cc|cc|}
  \toprule
   Number & \multicolumn{2}{c|}{Gaussian noise (std 30)}& \multicolumn{2}{c|}{Gaussian blur (sigma 4)}\\
     of blocks & POL &  independent & POL &  independent   \\
    \midrule
    1 &  23.26~\stdminus{0.15} & 23.26~\stdminus{0.15} &  18.61~\stdminus{0.13} & 18.61~\stdminus{0.13} \\
    2 &  23.28~\stdminus{0.07} & 23.21~\stdminus{0.05}&  18.48~\stdminus{0.25} & 18.64~\stdminus{0.21}\\
    4 &  24.41~\stdminus{0.27} &  23.16~\stdminus{0.09}&  19.19~\stdminus{0.04} & 19.17~\stdminus{0.51}\\
    8 &  23.91~\stdminus{0.23} & 22.27~\stdminus{0.02} &  18.95~\stdminus{0.14} & 19.33~\stdminus{0.29}\\
    12 &  23.91~\stdminus{0.10} & 22.48~\stdminus{0.32} &  19.70~\stdminus{0.21} & 18.76~\stdminus{0.09}\\
    16 &  23.88~\stdminus{0.14} & 22.54~\stdminus{0.48} &  19.00~\stdminus{0.30} & 18.11~\stdminus{0.41}\\
    \bottomrule
\end{tabular}
}
\end{table}

\paragraph{Choice of the maximum number of compositions.}

Table~\ref{tab:fusion_niqe_psnr_urban100_maxrep} compares the PSNR obtained for denoising and deblurring and NIQE for deblocking task on Urban-100~\cite{Huang2015Urban100}. 
We report more results and standard deviations compared to the main paper. 
Note that these tasks work well with a relatively low maximum of iterations, in contrast to style transfer image-to-image translations, which require more complex functions. 

\begin{table*}
\caption{\label{tab:fusion_niqe_psnr_urban100_maxrep}
Comparison between different maximum number of compositions. 
We report the most adapted metric on Urban-100: PSNR for the Gaussian noise and blur, NIQE for  deblocking. 
}
\centering
{
\small
\begin{tabular}{|c|c|c|c|}
  \toprule
    &\multicolumn{2}{c|}{PSNR (higher=better)} & \multicolumn{1}{c|}{NIQE (lower=better)}\\
     $\maxcomp$ & Gaussian noise (std 30) & Gaussian blur ($\sigma$=4) &  JPEG (quality=25)  \\
    \midrule
    1 &23.26 \stdminus{0.15} &  18.61 \stdminus{0.13}  &  10.17\stdminus{0.55} \\
    2  & 23.28 \stdminus{0.07} &  18.48 \stdminus{0.25} &  10.78 \stdminus{0.39}  \\
    3 &  23.89 \stdminus{0.07} &  19.13 \stdminus{0.08} &  10.57 \stdminus{0.21}  \\
    4 &  \textbf{24.41 \stdminus{0.27}} &  19.19 \stdminus{0.04}   &  10.43 \stdminus{0.25}  \\
    5 &  23.79 \stdminus{0.59} &  19.06 \stdminus{0.21} &  10.65 \stdminus{0.52}  \\
    6 &  23.80  \stdminus{0.31} &  19.09 \stdminus{0.08} & 10.42 \stdminus{0.19}  \\
    7 &  23.64 \stdminus{0.27} &  19.12 \stdminus{0.13}  &  10.93 \stdminus{0.61}  \\
    8 &  23.91 \stdminus{0.23} &  18.95 \stdminus{0.14}   &  10.34 \stdminus{0.61}  \\
    12 &  23.91 \stdminus{0.10} &  \textbf{19.70~\stdminus{0.21}}  &  9.74 \stdminus{0.41}  \\
    16 &  23.88 \stdminus{0.14} &  19.00~\stdminus{0.30}  &  8.51 \stdminus{0.19}  \\
    17 &  23.97 \stdminus{0.17} &  18.83 \stdminus{0.21}   &  8.36 \stdminus{0.32}   \\
    18 &  24.17 \stdminus{0.18} &  18.97 \stdminus{0.13}  &  7.49 \stdminus{0.63}   \\
    24  &  23.83 \stdminus{0.20} &  18.56 \stdminus{0.16}  &  \textbf{7.22 \stdminus{0.45}}  \\
    27 &  23.62 \stdminus{0.02} &  18.48 \stdminus{0.36}  & 7.25 \stdminus{0.36}  \\
    30 &   23.45 \stdminus{0.10} & 19.09 \stdminus{0.31}  &  8.15 \stdminus{0.64} \\
    \bottomrule
\end{tabular}
}
\end{table*}

\paragraph{Composition step.} 
Table~\ref{tab:psn_urban100_step} compares different choices for the number of steps of augmentation associated with the number of compositions $\maxcomp$. 
As we can see, taking too large steps tends to affect performance,
it is better to ramp up the number of compositions quickly during the warm-up phase.
\begin{table*}
\caption{\label{tab:psn_urban100_step}
PSNR on Urban-100~\cite{Huang2015Urban100} -- Gaussian noise (std=30).
Comparison between augmentation steps during the warm-up phase.
The augmentation step is to the number of epochs performed with the same number of compositions (1 epoch corresponds to 800 backward passes). 
}
\smallskip 
\centering
\scalebox{1.0}{
\small
\begin{tabular}{|c|c|c|c|c|c|}
    \toprule
     $\maxcomp$ & \multicolumn{5}{c|}{augmentation step}\\
       &1 & 2 & 4 & 8 & 16 \\
    \midrule
       4 & 24.41 \stdminus{0.27}     & 24.44 \stdminus{0.11}   &  23.62  \stdminus{0.16}   &  23.93\stdminus{0.10}   &   23.22 \stdminus{0.07}\\
     \midrule
       16 & 23.88 \stdminus{0.14} & 23.97  \stdminus{0.22} &  23.57 \stdminus{0.06} &  23.52 \stdminus{0.1}     & 23.19 \stdminus{0.02}  \\
    \midrule
       30  &  23.45 \stdminus{0.10}   &  23.77 \stdminus{0.14} & 23.95 \stdminus{0.17} & 23.14 \stdminus{0.04}  &23.17 \stdminus{0.04}   \\
    \bottomrule
\end{tabular}
}
\vspace{-1em}
\end{table*}

\begin{figure}

\centering
\small
\vspace{1em}
\begin{tabular}{c@{\hspace{0.75em}}c}
  \toprule
     fixed $\maxcomp=30$ & Random $\maxcomp \in [\![20,30]\!]$ \\
     \midrule
     \includegraphics[width=0.48\linewidth]{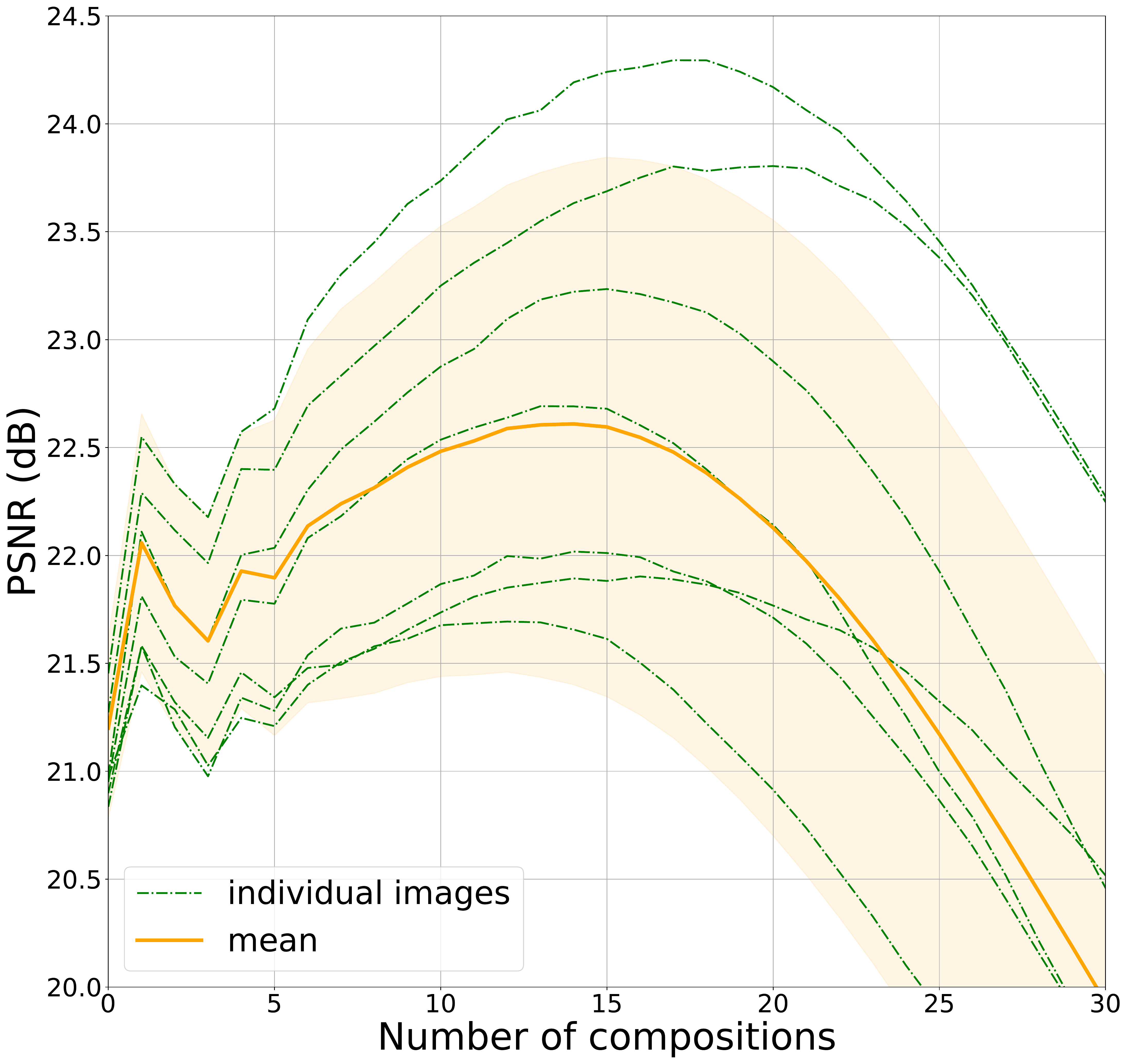}&
     \includegraphics[width=0.48\linewidth]{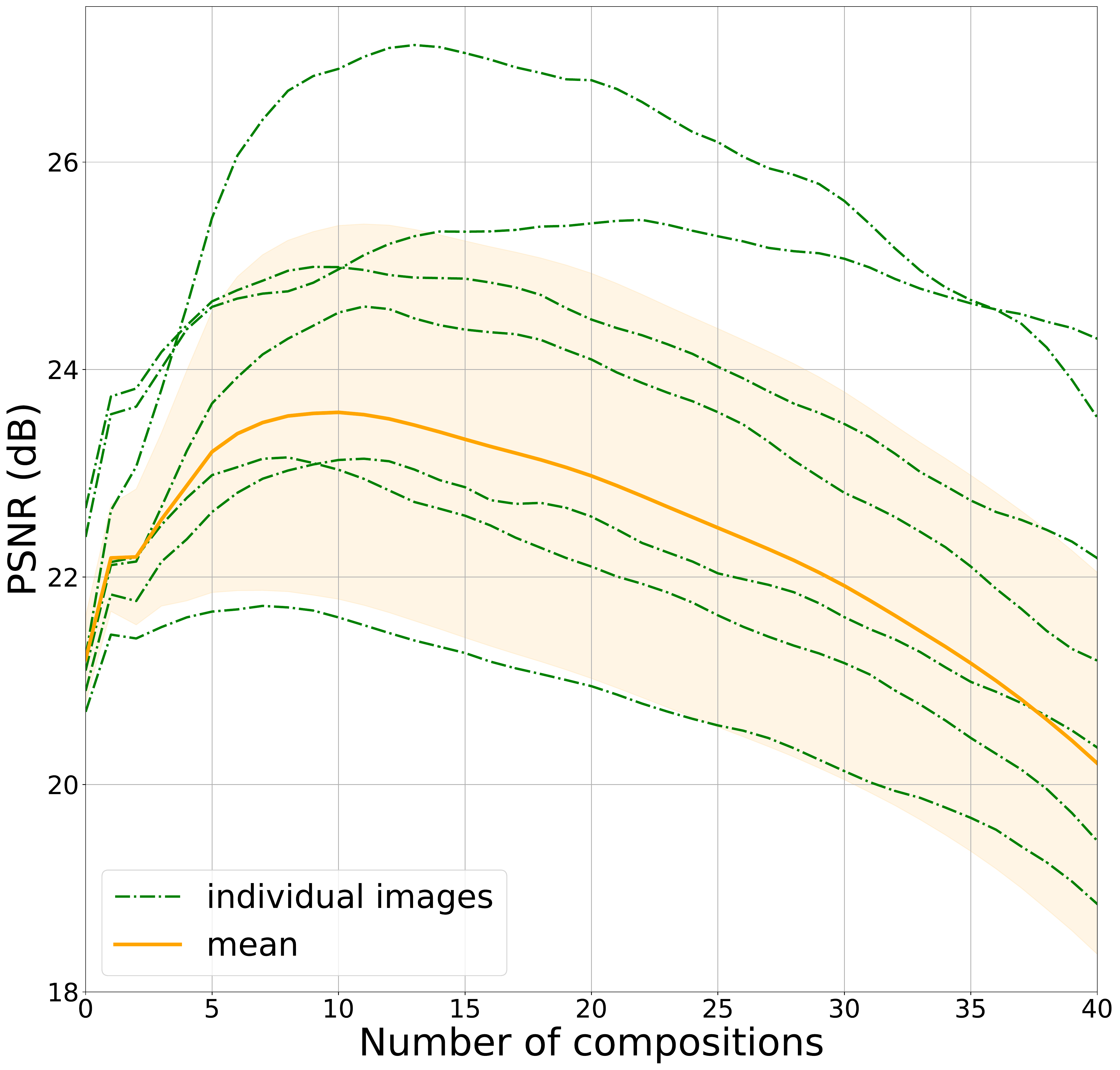} \\
    \bottomrule
\end{tabular}
\caption{\label{fig:ada_psnr}
 Evolution of the average PSNR as well as for different individual images according to the number of compositions $\maxcomptest$.
 The maximum number of compositions used for training is $\maxcomp=30$ and the Gaussian Noise standard deviation is 30.
}
\end{figure}

\paragraph{Comparison between randomised and fixed number of compositions~$\maxcomp$.} 
Figure~\ref{fig:ada_psnr} compares the trajectories of PSNRs as a function of the number of Powers-of-layers composition.
We get a better average performance if we randomly draw the maximum number of composition.

The different positions of the maxima in the adaptive case also suggests that it is necessary to adjust the amount of transformation to each image, as discussed below.

\section{Analysis of choices at inference time}
\label{app:inference}

\begin{figure*}
    \centering
    \includegraphics[width=.7\linewidth]{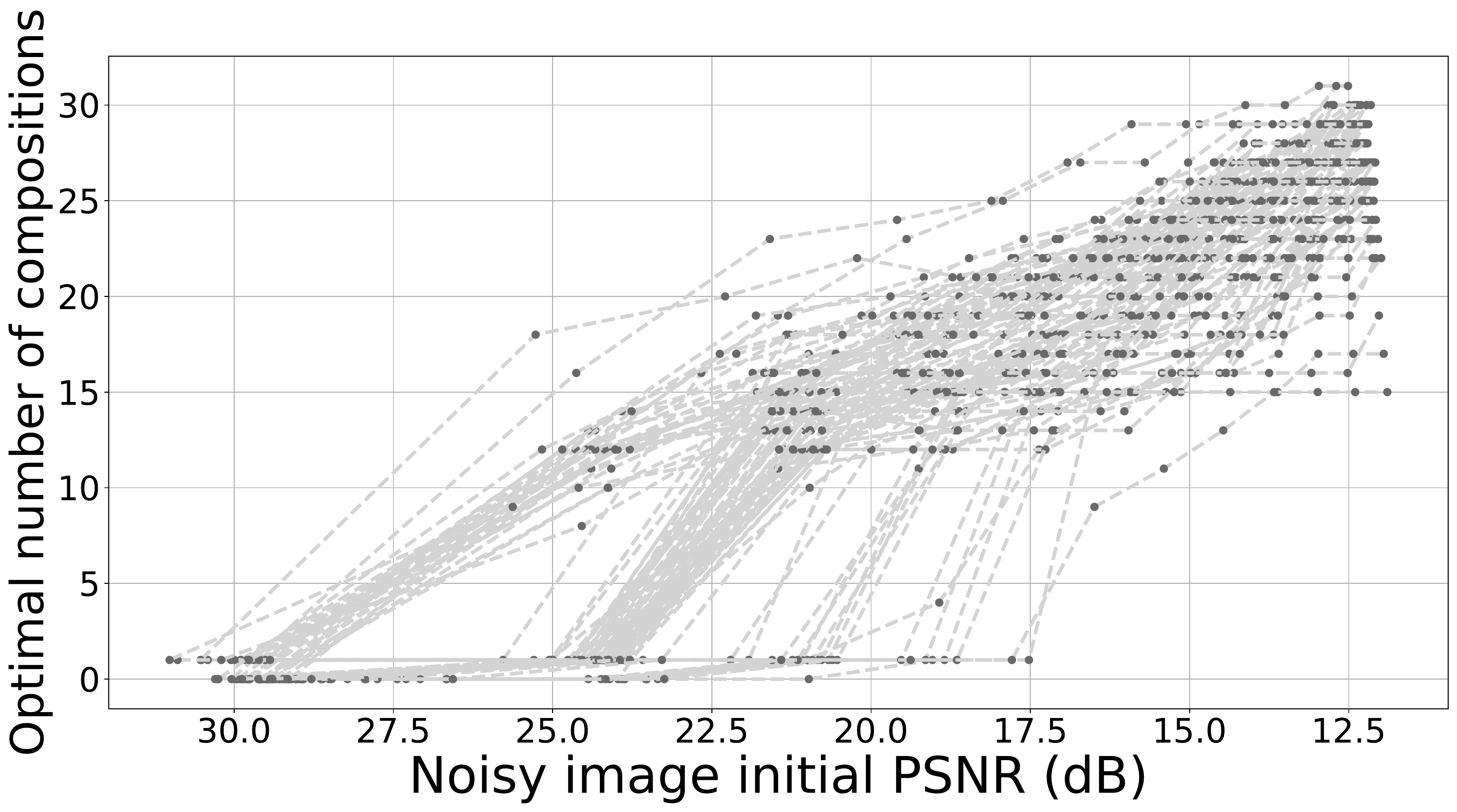}
    \caption{\label{fig:noise_complexity}
    Evolution of the optimal number of composition according to the noise. 
    The maximum number of compositions used for training is 30 and the Gaussian Noise standard deviation is 30. 
    }
\end{figure*}

\paragraph{Adjusting $\maxcomptest$ at inference time.}
Each image is more or less distant from the target domain, so we explore adapting the transformation to each image rather than applying a fixed transformation. 
For example, depending on the amount of noise, we may want to adjust the strength of the denoising. 
By modulating $\maxcomptest$ we can adapt the transformation to each image.
Figure~\ref{fig:noise_complexity} shows that the more noisy the input image is, the more we should compose to best denoise with Powers-of-Layers. 

\paragraph{Stopping criterion: fixed, discriminator, versus an Oracle.} 
At inference time the number of composition $\maxcomptest$ applied to each image can be set using several strategies.
We can choose to apply a constant number of composition or use the discriminator to choose the $\maxcomptest$ for which it gets the best response.

Table~\ref{tab:rand_comp} compares the performance of two strategies at test time for different random ranges at training time, and compare it to the upper bound achieved by an Oracle (i.e,. the performance attained when the optimal number of iteration is known for each image). 

With $\maxcomp=$30, we have chosen different ranges of the form $[\![d \times 30,30]\!]$ for $d\in \{100\%,75\%,66\%,50\%,33\%,25\%,0\%\}$.
The optimal range for debluring and denoising is with $d=66\%$

\begin{table*}
\caption{\label{tab:rand_comp}
Effect of setting the number of compositions $\maxcomp$ randomly on the PSNR on Urban-100 with two types of noise. 
We compare  
(Constant) a fixed $\maxcomptest=30$
and (Adaptive) value $\maxcomptest$ maximizing the target discriminator error for each image. 
Oracle: $\maxcomptest$ minimizing PSNR for each image.}
\smallskip
\centering
{
\small
\begin{tabular}{|c|ccc|ccc|}
  \toprule
     & \multicolumn{3}{c}{Gaussian Noise (std=30)} & \multicolumn{3}{c|}{Gaussian Blur ($\sigma$=4)} \\
     Random range  & Constant & Adaptive & Oracle & Constant & Adaptive & Oracle \\
    \midrule	
    $[\![0,30]\!]$ & 21.41\stdminus{3.33} & 22.11 \stdminus{0.56} & 23.86 \stdminus{0.27} &16.95\stdminus{1.29} & 16.16 \stdminus{1.33} & 19.43 \stdminus{0.09}\\
    $[\![7,30]\!]$ & 22.56\stdminus{1.94}  & 22.84\stdminus{0.67} &  23.73 \stdminus{0.98} & 16.39\stdminus{2.00}  & 15.42\stdminus{2.26} &  19.58 \stdminus{0.14}\\
    $[\![10,30]\!]$ & 21.99\stdminus{1.20} & 23.14\stdminus{0.21} & 23.42\stdminus{0.32} &17.48\stdminus{1.56} & 18.10\stdminus{0.70} & \textbf{19.65}\stdminus{0.10}\\
    $[\![15,30]\!]$ & 21.41\stdminus{1.71}  & 23.47\stdminus{0.69} &  23.77 \stdminus{0.61} &18.14\stdminus{0.54}  & 18.81\stdminus{0.61} &  19.42\stdminus{0.07}\\
    $[\![20,30]\!]$ & 21.53\stdminus{1.90}  & \textbf{23.68\stdminus{0.17}} &   \textbf{23.99 \stdminus{0.06}} &18.58\stdminus{0.10}  &  \textbf{19.22} \stdminus{0.37} &   19.56 \stdminus{0.09} \\
    $[\![22,30]\!]$ & 21.16\stdminus{0.99}  & 23.08\stdminus{0.37} &  23.42 \stdminus{0.28} &17.55\stdminus{1.32}  & 19.06\stdminus{0.39} &  19.52 \stdminus{0.05} \\
    $[\![30,30]\!]$ & \textbf{23.45\stdminus{0.10}}  & 23.46\stdminus{0.40} &  23.81 \stdminus{0.48} &\textbf{19.09} \stdminus{0.31}  & 18.87\stdminus{0.41} &  19.49 \stdminus{0.08} \\
\bottomrule
\end{tabular}
}
\end{table*}

\section{Additional comparisons with CycleGAN and the Fader Network} 
\label{app:cyclegan}

Table~\ref{tab:psnr_urban100_comp_1} compares the results obtained with PoL and CycleGAN for different noise levels.
Table~\ref{tab:data_comp} compares the results obtained with PoL and CycleGAN for different amounts of data.
These numbers are the same as Figure~\ref{tab:psnr_urban100_comp}, with standard deviations.
In most cases, whether with different amounts of data or different noise, our method is better than CycleGAN.
This is mainly due to its smaller number of parameters and the flexibility brought by the adaptive criterion.

\begin{table*}
\caption{\label{tab:psnr_urban100_comp_1}
Comparison between our approach (PoL) and CycleGAN. 
We three tasks, all computed with the Urban-100 dataset: 
PSNR (higher is better) with different amount of Gaussian noise and Gaussian blur, and 
NIQE (lower is better) measured for different JPEG compression quality.
}
\smallskip
\centering
\scalebox{0.73}{
\small
\begin{tabular}{@{}|c|ccc|c|ccc|c|ccc|}
  \toprule
     Noise &  Noisy & \multicolumn{2}{c}{Denoising}  & Sigma & Blur & \multicolumn{2 }{c|}{Debluring}& JPEG & JPEG &\multicolumn{2}{c|}{Deblocking}\\
        (std) &   images & CycleGAN & PoL &   Blur &  images & CycleGAN & PoL&   quality &  images & CycleGAN & PoL\\
    \midrule	
    15 & 24.9& 22.37\stdminus{0.19}  &\textbf{27.37} \stdminus{0.26} &
    2 & 21.58 & 20.37 \stdminus{0.26} & \textbf{22.14}  \stdminus{0.21}&
    15& 9.01  & 7.89\stdminus{1.14}  &7.90\stdminus{0.32}\\
    30 & 19.2   & 21.93\stdminus{0.04} & \textbf{23.68} \stdminus{0.17} & 
    4 & 19.20  & 18.55 \stdminus{0.37} & \textbf{19.22} \stdminus{0.37}&
    25 &  8.94 & 7.45\stdminus{0.63}  & \textbf{7.10}\stdminus{0.58} \\
    50 & 15.2  & 21.57\stdminus{0.04} & \textbf{ 22.52} \stdminus{0.37} &
    8 &  17.48  & 16.09 \stdminus{0.14} & \textbf{17.53} \stdminus{0.34}&
    30 & 8.90  & 7.46\stdminus{0.32}  & \textbf{6.76}\stdminus{0.83}\\
    70 & 12.7  & \textbf{21.02}\stdminus{0.17}& 20.94 \stdminus{0.24} &
    16 & 16.13  & 16.11 \stdminus{0.31} & \textbf{16.16} \stdminus{0.14} &
    50 & 8.99  & 6.96\stdminus{0.55}  &   \textbf{6.56}\stdminus{0.48}\\
    100 & 10.4  & \textbf{20.00}\stdminus{0.12} & 19.42\stdminus{0.22} & 
    24 & 15.50  & 12.88 \stdminus{0.24} & \textbf{15.48} \stdminus{0.07}&
    70 & 8.94 & 7.11\stdminus{0.21}  &   \textbf{6.81}\stdminus{0.39}\\
\bottomrule
\end{tabular}
}
\vspace{-1em}
\end{table*}

\begin{table*}
\caption{\label{tab:data_comp}
Comparison between CycleGAN and Power of layer on Urban-100~\cite{Huang2015Urban100} with different amount of training data. We use PSNR to compare methods for Gaussian noise and Gaussian blur.
}
\smallskip
\centering
\small
\begin{tabular}{|c|cc|cc|}
  \toprule
   Number of    data & \multicolumn{2}{c|}{Denoising (std=30)} & \multicolumn{2}{c|}{Debluring (sigma=4)}\\
      training images & CycleGAN  & PoL & CycleGAN  & PoL  \\
    \midrule	
    1 & 
    14.50\stdminus{0.24} & 
    \textbf{22.27\stdminus{0.16}} & 
    14.36\stdminus{0.12} & 
     \textbf{18.68 \stdminus{0.27}}\\
    5 &
    16.57\stdminus{0.04} & 
     \textbf{23.13\stdminus{0.18}} &
    16.72\stdminus{0.15} & 
     \textbf{18.76 \stdminus{0.23}} \\
    10 & 
    20.88\stdminus{0.36}& 
     \textbf{23.19\stdminus{0.58}} & 
    17.03 \stdminus{0.07} & 
     \textbf{18.82 \stdminus{0.56}}\\
    100 & 
    21.03\stdminus{0.76}  & 
     \textbf{23.39\stdminus{0.33}}&
    18.17 \stdminus{0.15} & 
     \textbf{18.97 \stdminus{0.23}}\\
    400 & 
    21.78\stdminus{0.12} & 
     \textbf{23.60 \stdminus{0.32}} & 
    18.21\stdminus{0.11} & 
     \textbf{19.01 \stdminus{0.31}} \\ 
    800 & 
    21.93\stdminus{0.04} &
     \textbf{23.88 \stdminus{0.14}} & 
    18.55 \stdminus{0.37} & 
     \textbf{19.22 \stdminus{0.37}} \\
\bottomrule
\end{tabular}
\end{table*}

\paragraph{Experiments on transformation  adjustment}

As baseline  we use the Fader network~\cite{lample2017fader} for transformation adjustment.
The Fader Network is a neural network composed of an encoder and a decoder, 
for which it it is possible to modulate a transformation. This is done by removing the factors of variations related to this transformation in the latent space resulting from the encoder, and in turn by choosing the factors to be added to the embedding going into the decoder.

To interpolate between domain $\Ac$ and domain $\Bc$, the Fader network has a latent representation where the attributes relative to each domain have been disentangled. 
The  Fader network has been applied to faces, for instance to add glasses on a face, to age a person, etc.
We observe experimentally with smaller datasets, where the variability from one image to another is larger than with faces, that the Fader's results are not as good. 

In contrast, our approach, like CycleGAN, does not have limitations incurred by a latent space disentanglement because it exploits a cyclic loss.

\begin{figure*}
\centering
\newcommand{\fadercmp}[1]{\includegraphics[width=0.13\linewidth]{images/Fader/#1}}
\centering
\small
\begin{tabular}{c|c|ccccc}
 \multicolumn{2}{c}{} &\multicolumn{5}{c}{\large Transformation rate} \\
 Original & & 0\%  & 25\%  & 50\% & 75\% & 100\%  \\
 \midrule
 \fadercmp{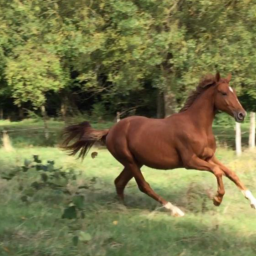} &
 \rotatebox{90}{Fader,  3 layers} &
  \fadercmp{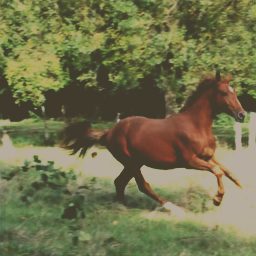} & \fadercmp{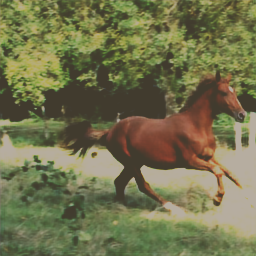} & \fadercmp{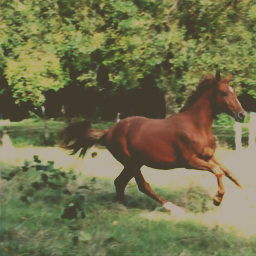} & \fadercmp{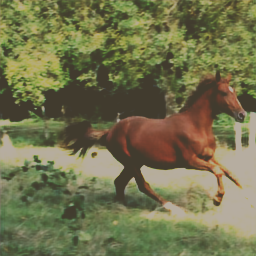} & \fadercmp{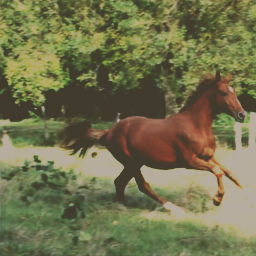}\\

 \fadercmp{small/org_images_0.png} &
 \rotatebox{90}{Fader, 6 layers} &
  \fadercmp{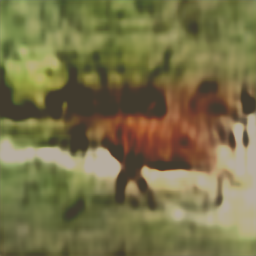} & \fadercmp{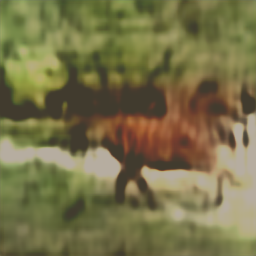} & \fadercmp{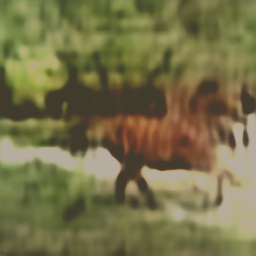} & \fadercmp{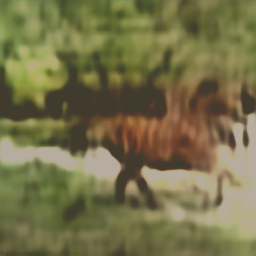} & \fadercmp{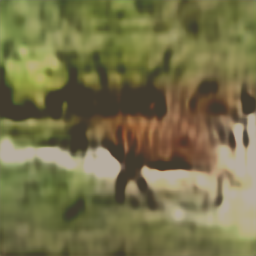}\\
 
 \midrule

  \fadercmp{small/org_images_0.png} &
   \rotatebox{90}{PoL} &
 \fadercmp{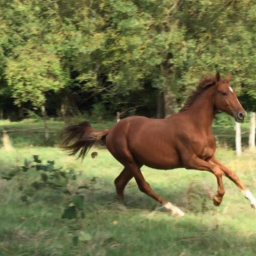} & \fadercmp{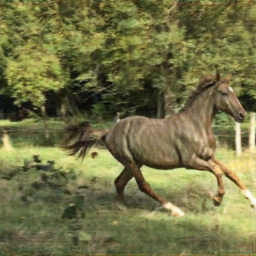} & \fadercmp{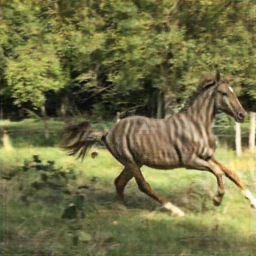} & \fadercmp{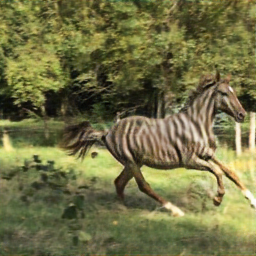} & \fadercmp{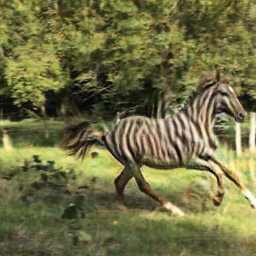}\\
 
\end{tabular}
\caption{\label{fig:fader_horse}
Visual comparison between Fader networks~\cite{lample2017fader} and our power-of-layers on the task Horse to Zebra. 
}
\vspace{-1em}
\end{figure*}

Figure~\ref{fig:fader_horse} shows the results obtained with the Fader network and with our method on the Horse$\rightarrow$Zebra transformation adjustment.
The Fader network is unable to significantly transform the source when the network is too shallow (3 layers) and destroys the image when it is deep (6 layers). 
In contrast, Powers-of-layers convincingly hybridizes a horse and a zebra.
In terms of FID for the Horse to Zebra task, the Fader network is significantly worse: it obtains a FID greater than 163.0 in the both case against 53.0 for our method (lower is better).

\section{Additional results: visualizations of transform modulation and high resolution}
\label{app:visu}

\paragraph{Progressive results}
Figure~\ref{tab:continuousappendix} shows the progressive transformations obtained on different tasks. 
It shows that progressive transformations are realistic for most tasks.

\def \mysp {\hspace{5.4pt}}

\begin{figure*}
\centering
\scalebox{0.75}{
\small
\begin{tabular}{c|@{}c@{\mysp}c@{\mysp}c@{\mysp}c@{\mysp}c@{\mysp}c@{\mysp}c@{\mysp}c@{\mysp}c@{}}
Domain / Composition  & 0 & 5 & 10 & 15 & 20 & 25 & 30 & 35 & 40\\
 Horse $\rightarrow$ Zebra & \includegraphics[width=0.11\linewidth]{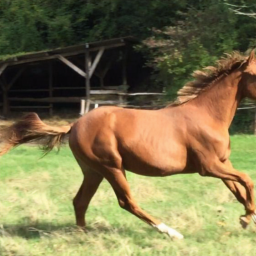} & \includegraphics[width=0.11\linewidth]{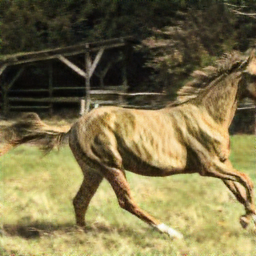} &
\includegraphics[width=0.11\linewidth]{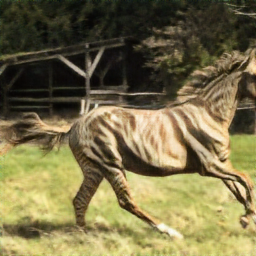} &
\includegraphics[width=0.11\linewidth]{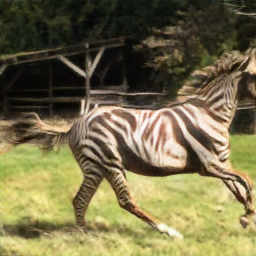} &
\includegraphics[width=0.11\linewidth]{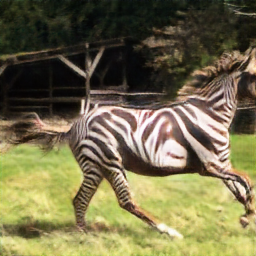} &
\includegraphics[width=0.11\linewidth]{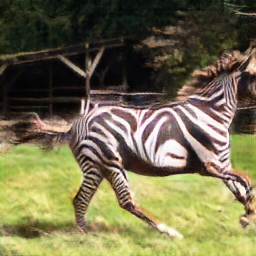} &
\includegraphics[width=0.11\linewidth]{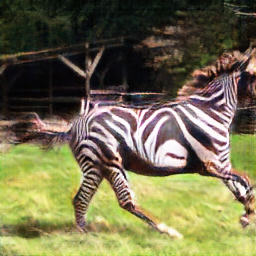} &
\includegraphics[width=0.11\linewidth]{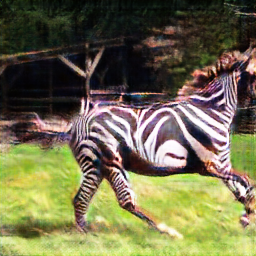} &
\includegraphics[width=0.11\linewidth]{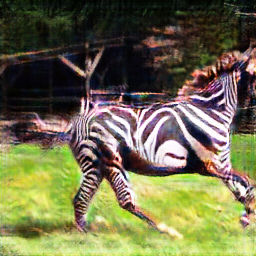} \\

Zebra $\rightarrow$ Horse &\includegraphics[width=0.11\linewidth]{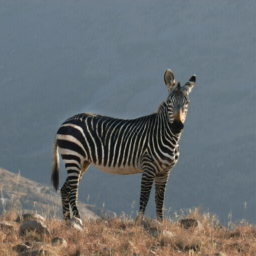} & \includegraphics[width=0.11\linewidth]{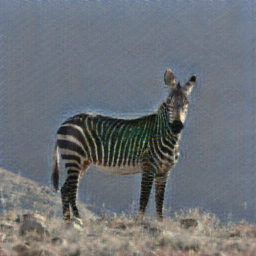} &
\includegraphics[width=0.11\linewidth]{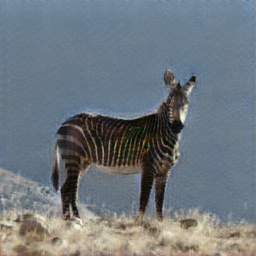} &
\includegraphics[width=0.11\linewidth]{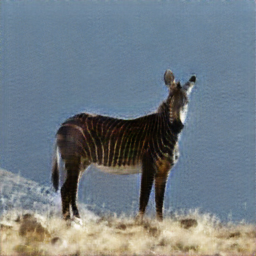} &
\includegraphics[width=0.11\linewidth]{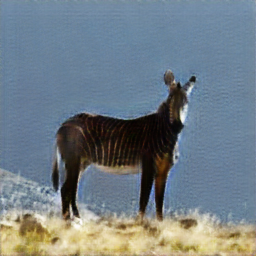} &
\includegraphics[width=0.11\linewidth]{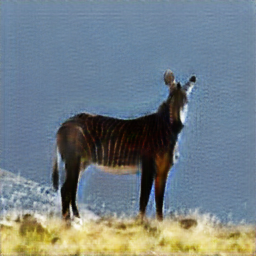} &
\includegraphics[width=0.11\linewidth]{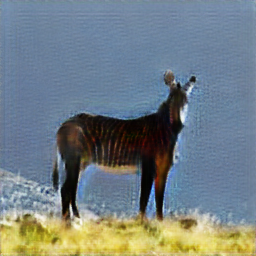} &
\includegraphics[width=0.11\linewidth]{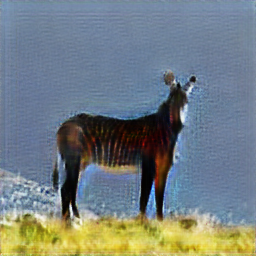} &
\includegraphics[width=0.11\linewidth]{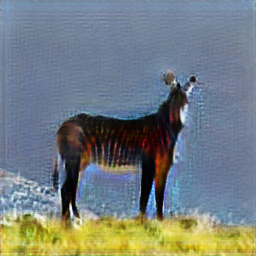} \\

Summer $\rightarrow$ Winter &\includegraphics[width=0.11\linewidth]{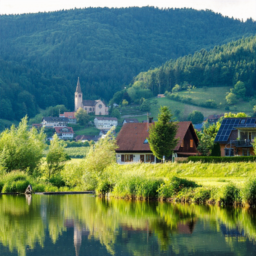} & \includegraphics[width=0.11\linewidth]{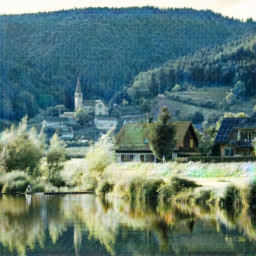} &
\includegraphics[width=0.11\linewidth]{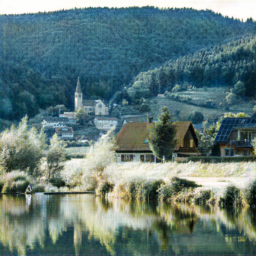} &
\includegraphics[width=0.11\linewidth]{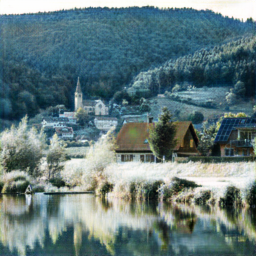} &
\includegraphics[width=0.11\linewidth]{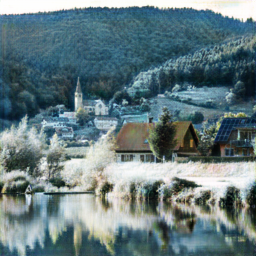} &
\includegraphics[width=0.11\linewidth]{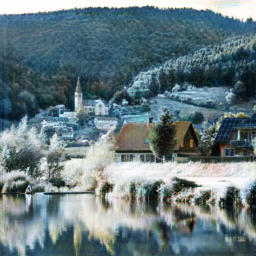} &
\includegraphics[width=0.11\linewidth]{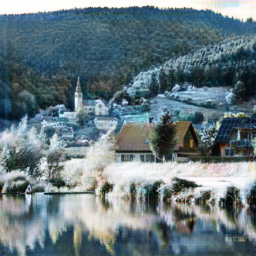} &
\includegraphics[width=0.11\linewidth]{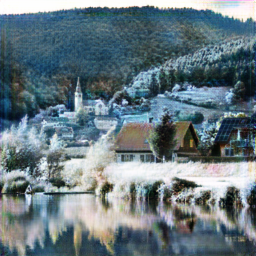} &
\includegraphics[width=0.11\linewidth]{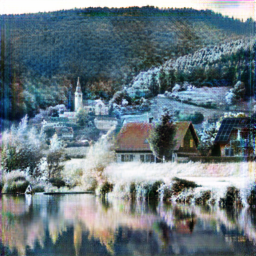} \\

Winter $\rightarrow$ Summer &\includegraphics[width=0.11\linewidth]{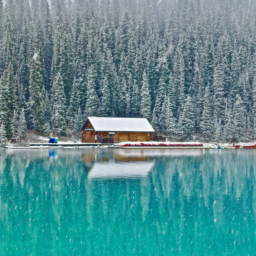} & \includegraphics[width=0.11\linewidth]{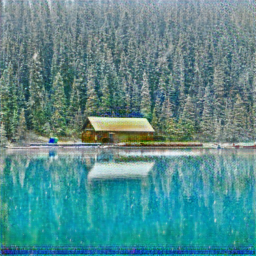} &
\includegraphics[width=0.11\linewidth]{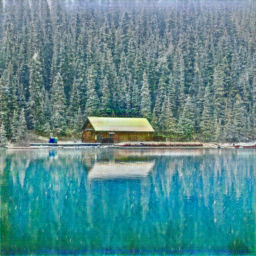} &
\includegraphics[width=0.11\linewidth]{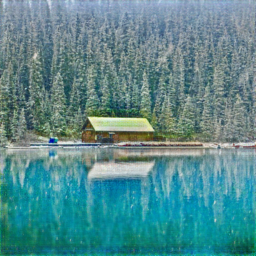} &
\includegraphics[width=0.11\linewidth]{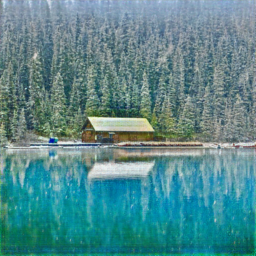} &
\includegraphics[width=0.11\linewidth]{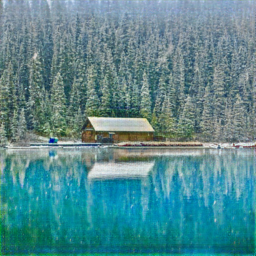} &
\includegraphics[width=0.11\linewidth]{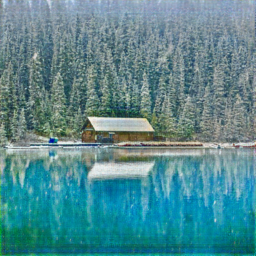} &
\includegraphics[width=0.11\linewidth]{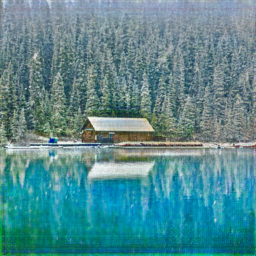} &
\includegraphics[width=0.11\linewidth]{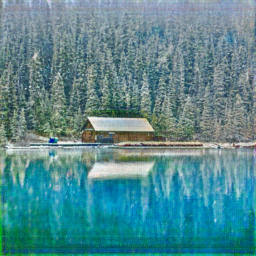} \\

Photo $\rightarrow$ Monet &\includegraphics[width=0.11\linewidth]{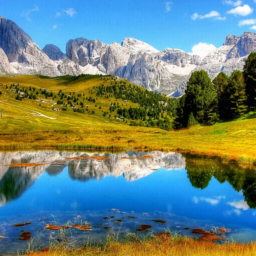} & \includegraphics[width=0.11\linewidth]{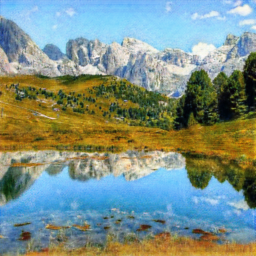} &
\includegraphics[width=0.11\linewidth]{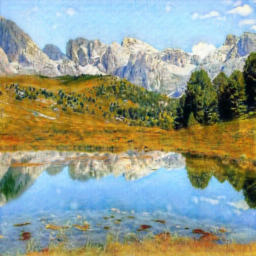} &
\includegraphics[width=0.11\linewidth]{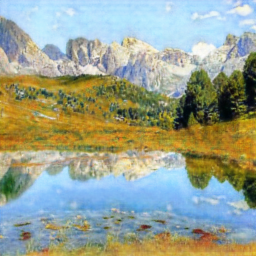} &
\includegraphics[width=0.11\linewidth]{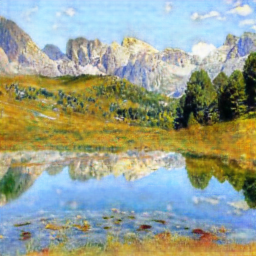} &
\includegraphics[width=0.11\linewidth]{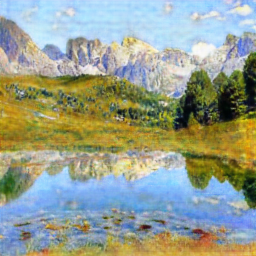} &
\includegraphics[width=0.11\linewidth]{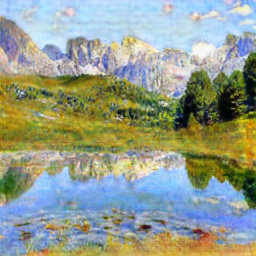} &
\includegraphics[width=0.11\linewidth]{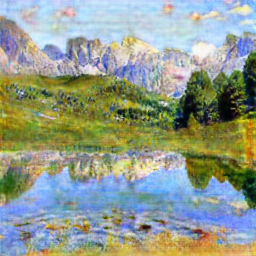} &
\includegraphics[width=0.11\linewidth]{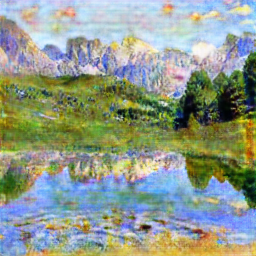} \\

Monet $\rightarrow$ photo & \includegraphics[width=0.11\linewidth]{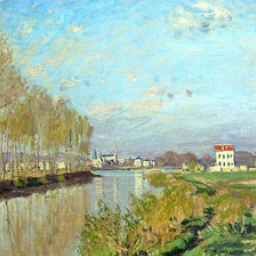} & \includegraphics[width=0.11\linewidth]{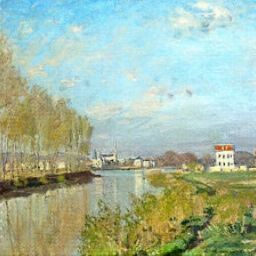} &
\includegraphics[width=0.11\linewidth]{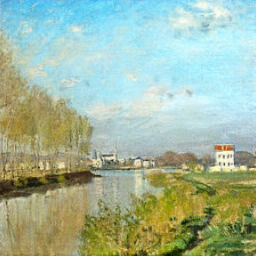} &
\includegraphics[width=0.11\linewidth]{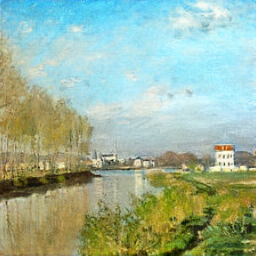} &
\includegraphics[width=0.11\linewidth]{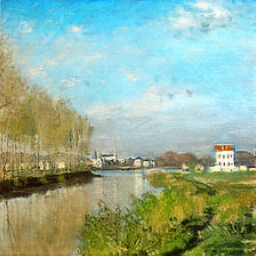} &
\includegraphics[width=0.11\linewidth]{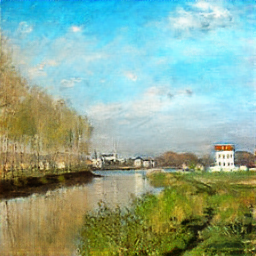} &
\includegraphics[width=0.11\linewidth]{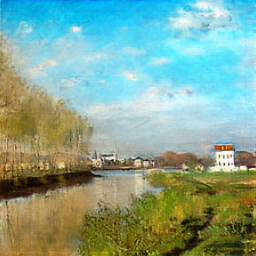} &
\includegraphics[width=0.11\linewidth]{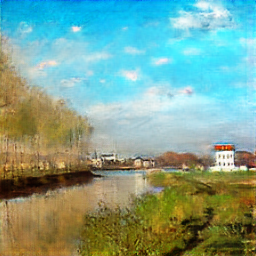} &
\includegraphics[width=0.11\linewidth]{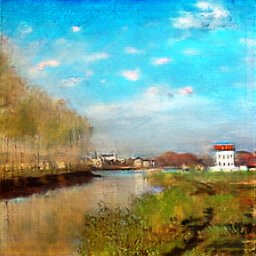} \\
\end{tabular}
}
\caption{\label{tab:continuousappendix}
Illustration of the different results obtained along the iterations of the recurrent block, during the transformation of a horse into a zebra.
The first image is the original image, then each image corresponds to 5 additional compositions of our method. Since our method was learned with a maximum of 30 compositions the last two images are extrapolations.
}
\vspace{-1em}
\end{figure*}

\begin{figure*}
\centering
\small
\scalebox{0.9}{
\begin{tabular}{ccc}
\multicolumn{3}{c}{Powers-of-Layers Transformations}\\
 Photo $\rightarrow$ Van Gogh & Photo $\rightarrow$ Monet  & Horse $\rightarrow$ Zebra \\
 \includegraphics[width=0.35\linewidth]{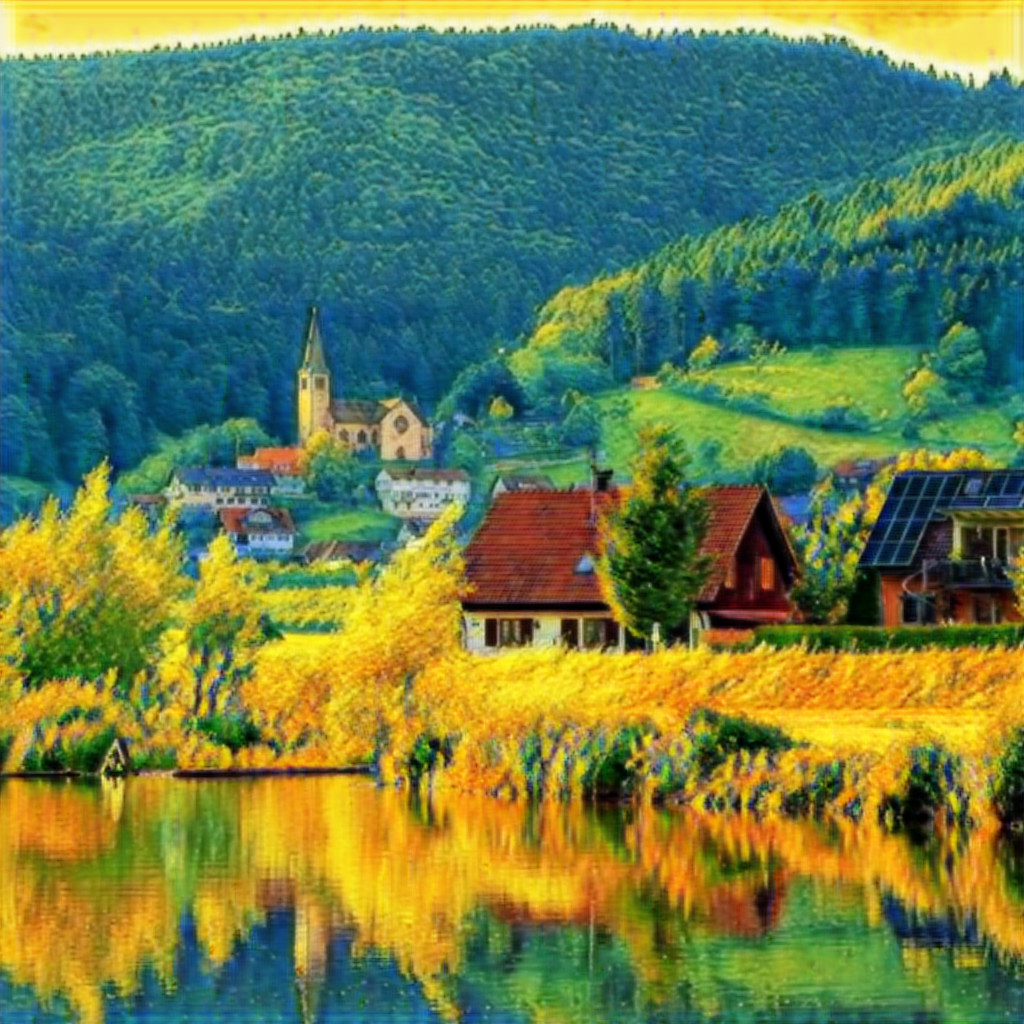}&
 \includegraphics[width=0.35\linewidth]{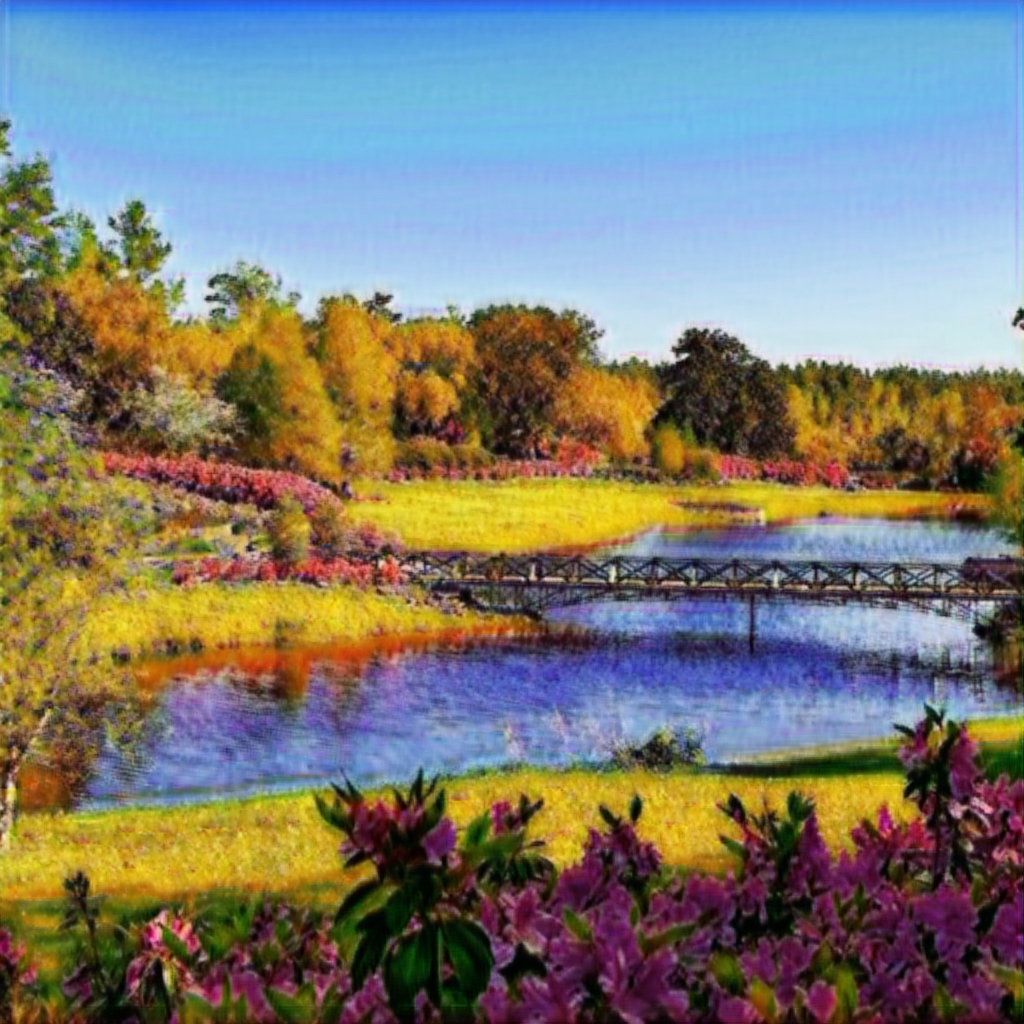} &
   \includegraphics[width=0.35\linewidth]{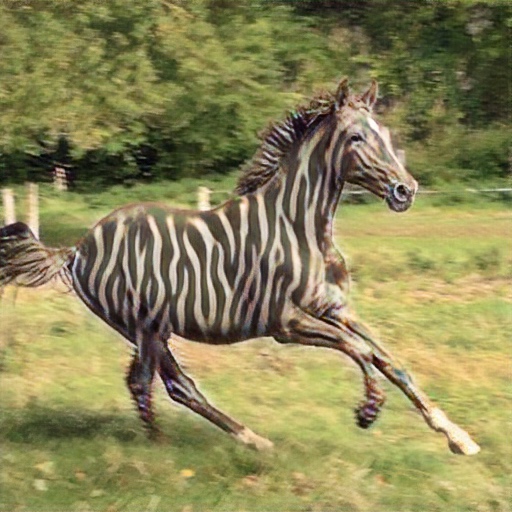}\\
  \includegraphics[width=0.35\linewidth]{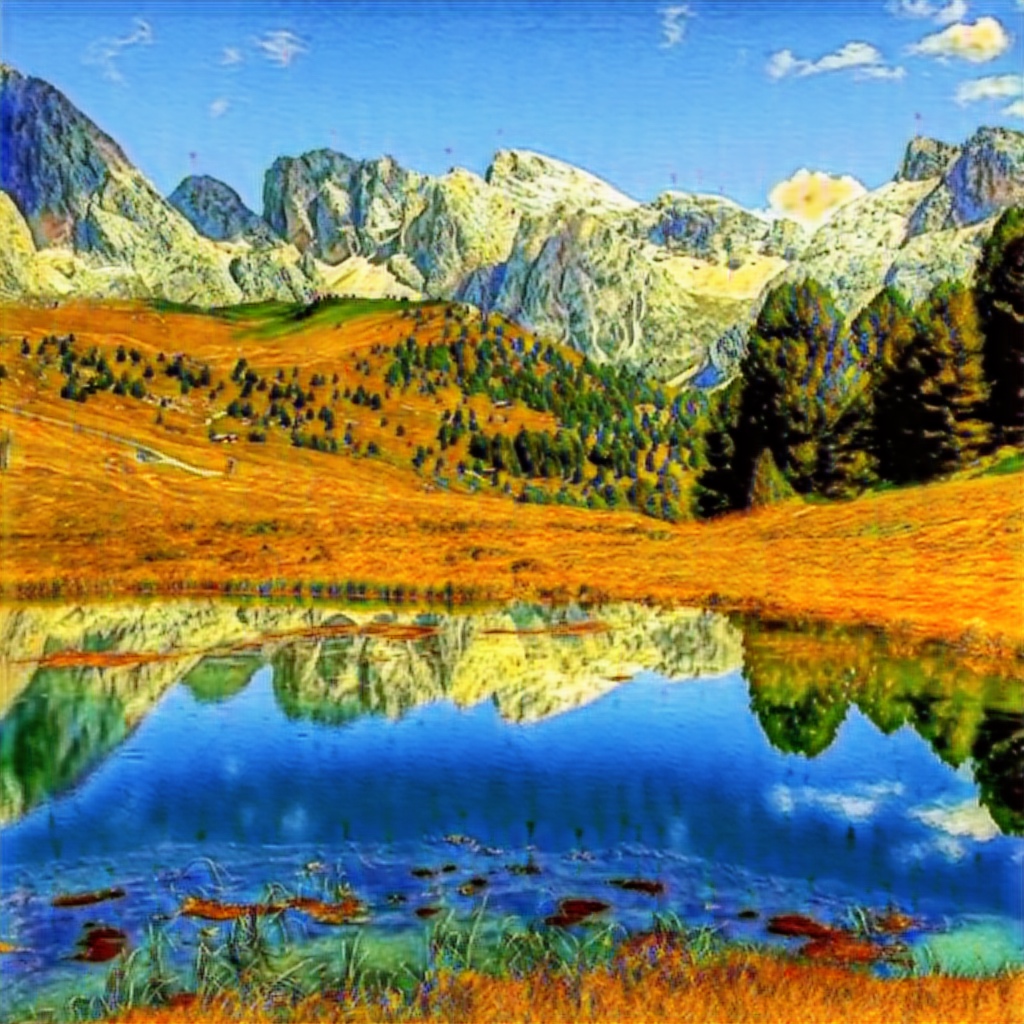}
  &\includegraphics[width=0.35\linewidth]{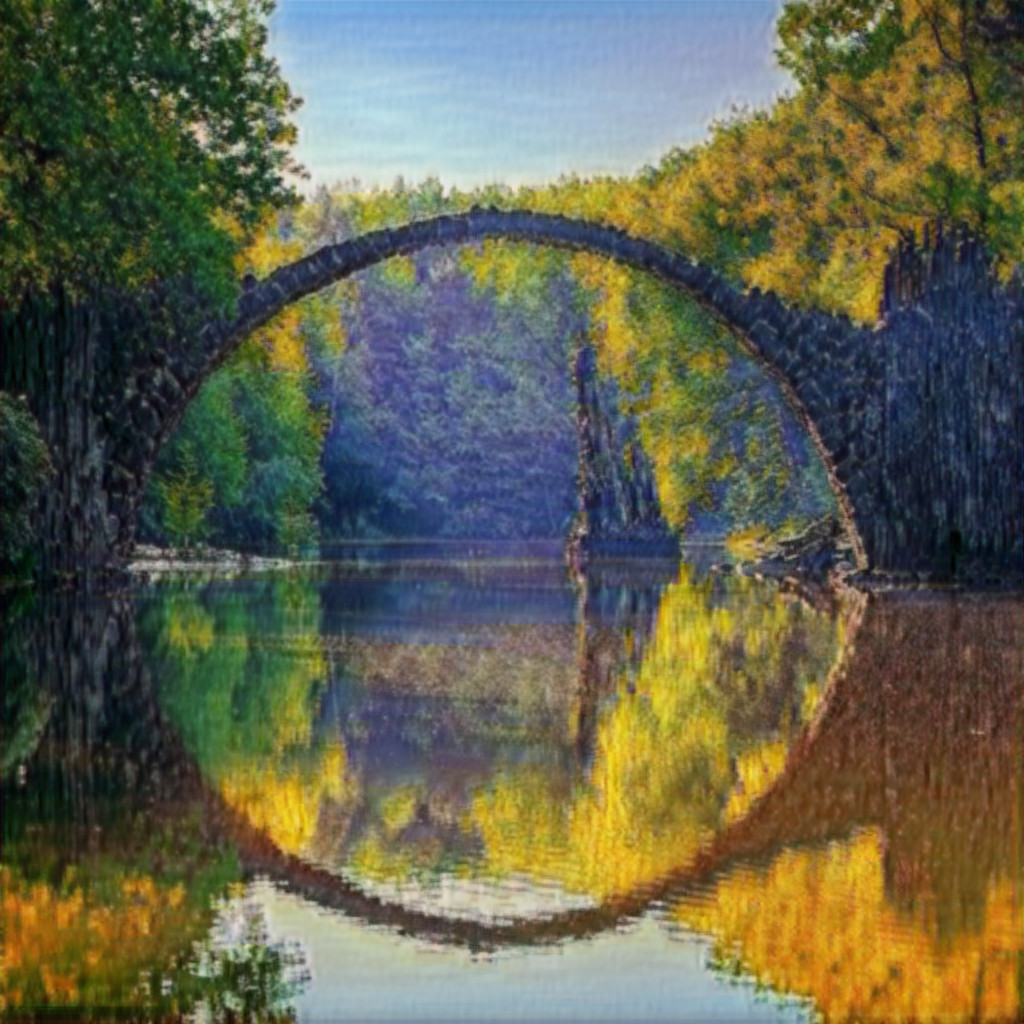} &
  \includegraphics[width=0.35\linewidth]{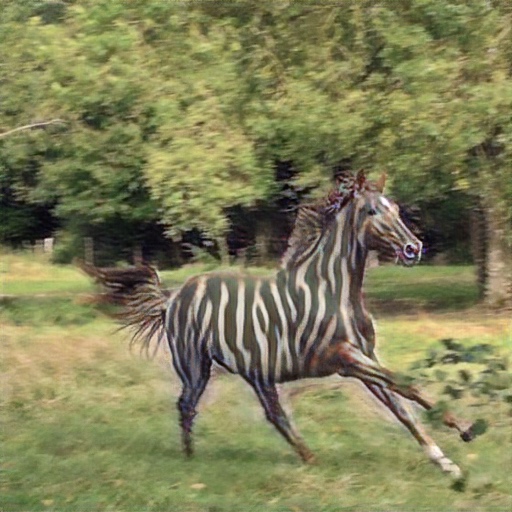}\\[0.1cm]
  \hline\\[0.1cm]
  \multicolumn{3}{c}{Original images}\\
   \includegraphics[width=0.35\linewidth]{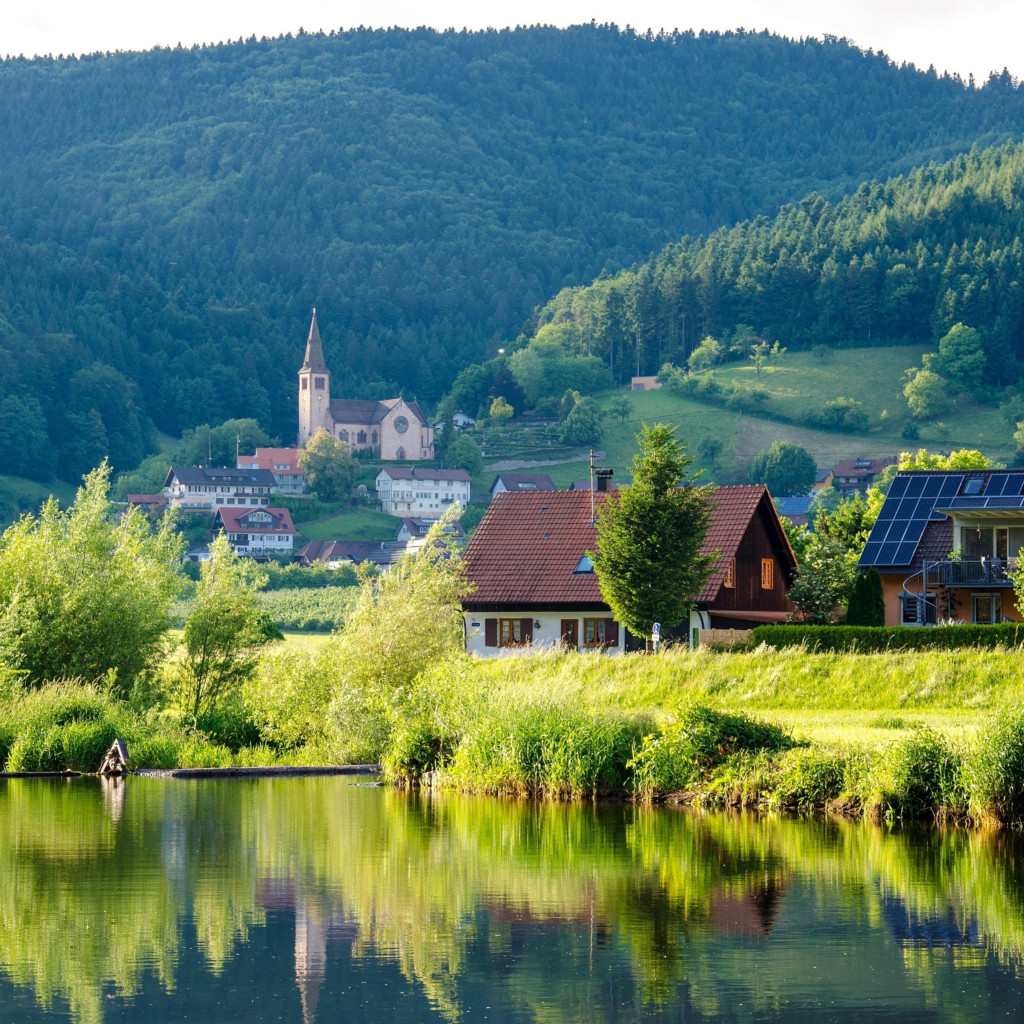}&
 \includegraphics[width=0.35\linewidth]{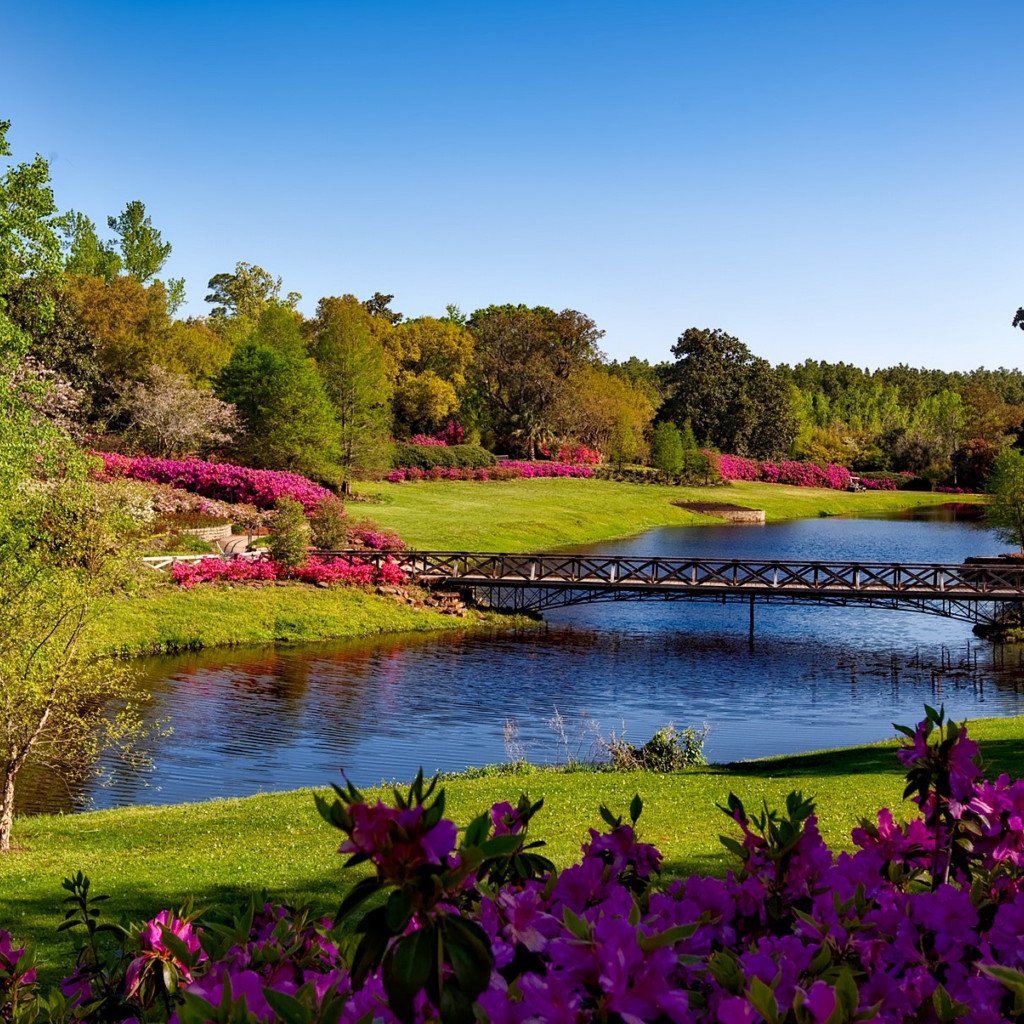}&
  
   \includegraphics[width=0.35\linewidth]{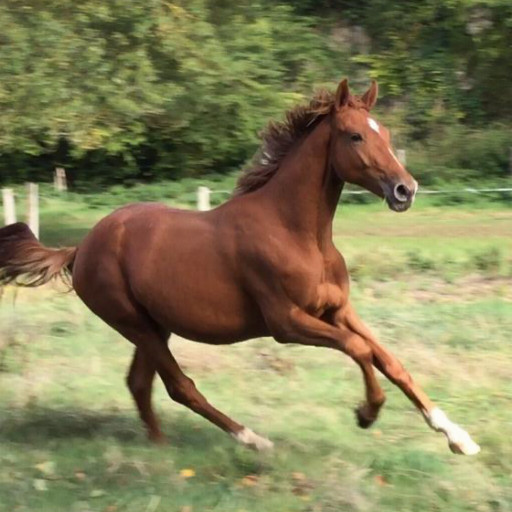}\\
  \includegraphics[width=0.35\linewidth]{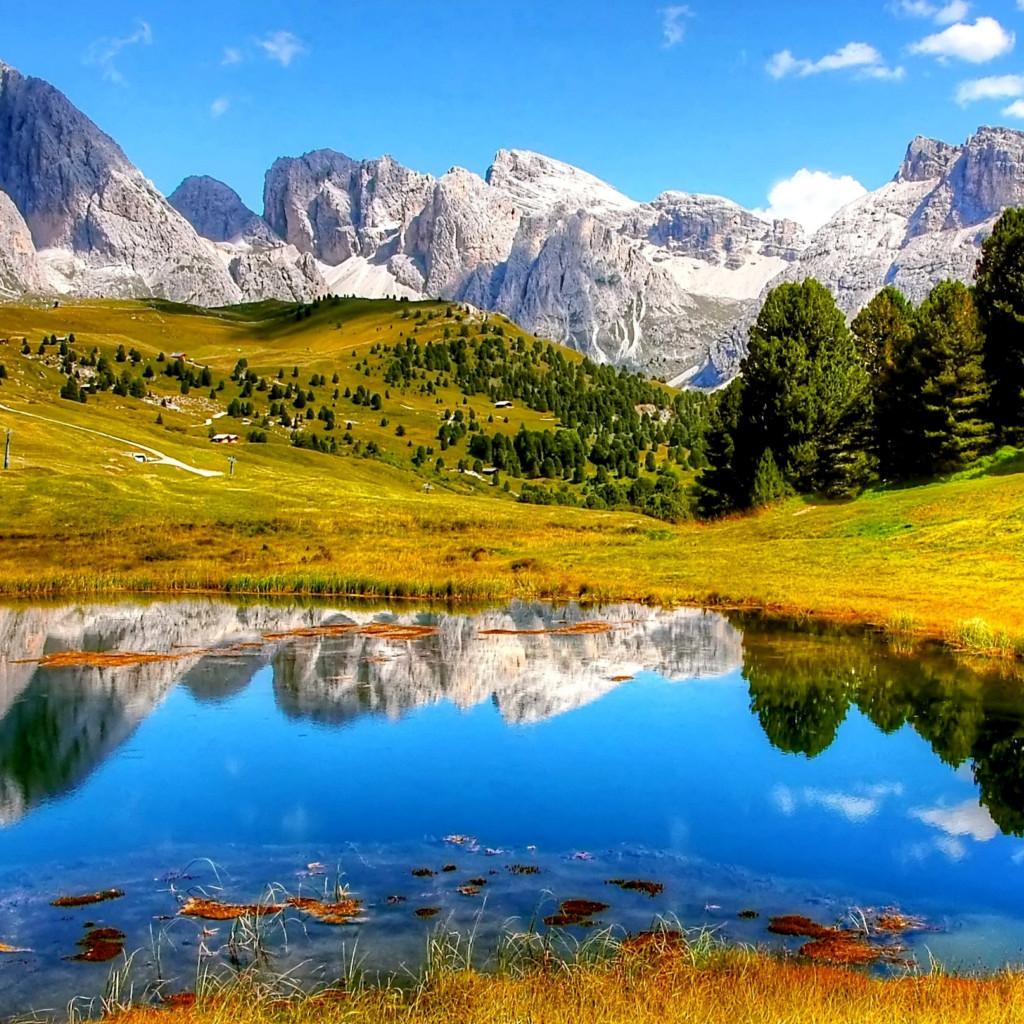}
  &\includegraphics[width=0.35\linewidth]{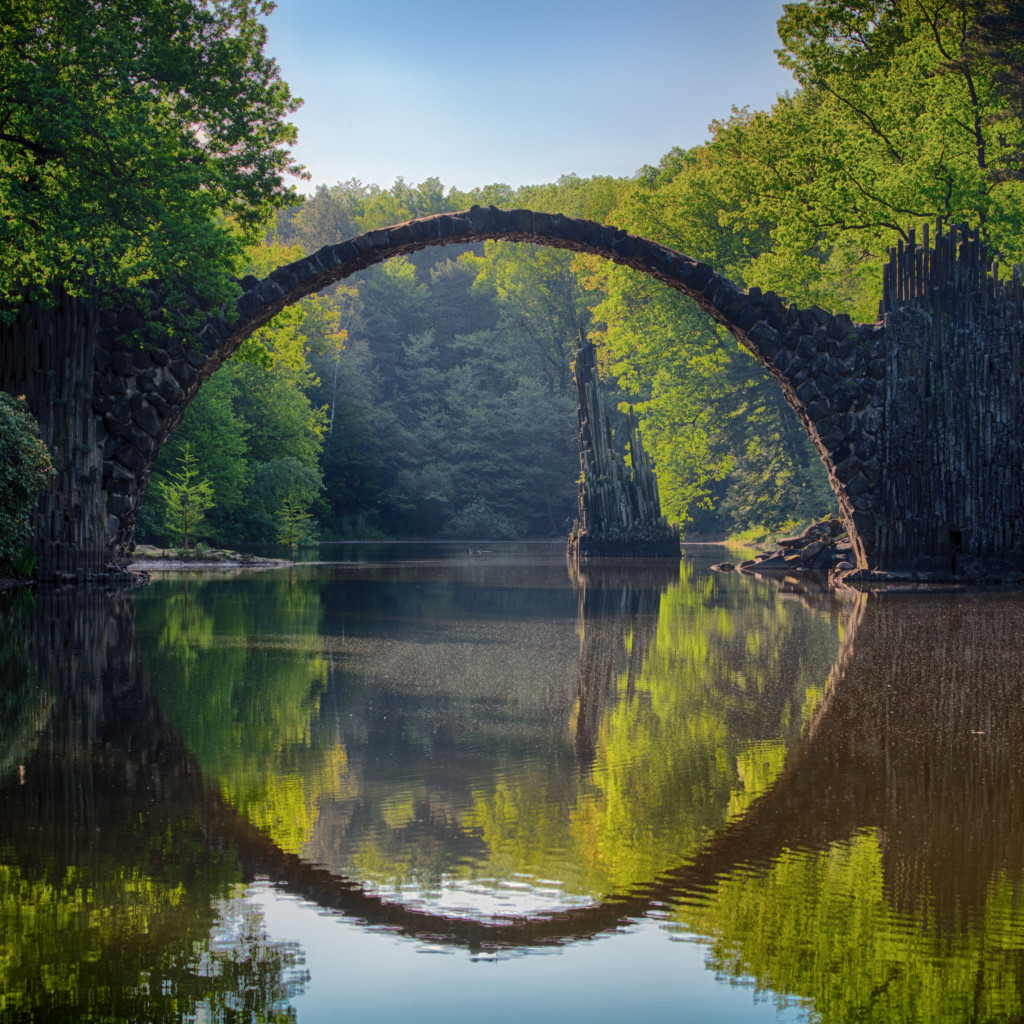}& 
  \includegraphics[width=0.35\linewidth]{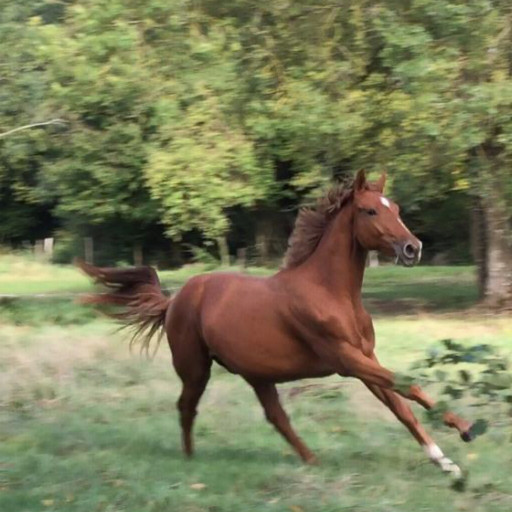}\\
\end{tabular}
}
\caption{\label{fig:highres_app}
{\it Top:} Differents visuals results with high resolution image. {\it Bottom:} Original images}
\end{figure*}

\paragraph{Results in high resolution.}
Figure~\ref{fig:highres_app} shows the high-resolution results of the section~\ref{sec:experiments}, along with the original images.

\end{document}